\documentclass[lettersize,journal]{IEEEtran}
\usepackage{amsmath,amsfonts}
\usepackage{algorithmic}
\usepackage{algorithm}
\usepackage{array}
\usepackage[caption=false,font=normalsize,labelfont=sf,textfont=sf]{subfig}
\usepackage{textcomp}
\usepackage{stfloats}
\usepackage{url}
\usepackage{verbatim}
\usepackage{graphicx}
\usepackage{cite}
\usepackage{xcolor}
\usepackage{adjustbox}
\usepackage{booktabs}
\usepackage{hyperref}
\definecolor{darkgreen}{HTML}{00B050}
\definecolor{darkred}{HTML}{E10600}

\usepackage{caption} 
\usepackage{scalerel}
\usepackage{tikz}
\usetikzlibrary{svg.path}

\definecolor{orcidlogocol}{HTML}{A6CE39}
\tikzset{
    orcidlogo/.pic={
        \fill[orcidlogocol] svg{M256,128c0,70.7-57.3,128-128,128C57.3,256,0,198.7,0,128C0,57.3,57.3,0,128,0C198.7,0,256,57.3,256,128z};
        \fill[white] svg{M86.3,186.2H70.9V79.1h15.4v48.4V186.2z}
        svg{M108.9,79.1h41.6c39.6,0,57,28.3,57,53.6c0,27.5-21.5,53.6-56.8,53.6h-41.8V79.1z M124.3,172.4h24.5c34.9,0,42.9-26.5,42.9-39.7c0-21.5-13.7-39.7-43.7-39.7h-23.7V172.4z}
        svg{M88.7,56.8c0,5.5-4.5,10.1-10.1,10.1c-5.6,0-10.1-4.6-10.1-10.1c0-5.6,4.5-10.1,10.1-10.1C84.2,46.7,88.7,51.3,88.7,56.8z};
    }
}

\newcommand\orcidicon[1]{\href{https://orcid.org/#1}{\mbox{\scalerel*{
                \begin{tikzpicture}[yscale=-1,transform shape]
                \pic{orcidlogo};
                \end{tikzpicture}
            }{|}}}}

\newcommand{\ourframework}{\textsc{SKG2Data}}

\newcommand{\testdata}{\textsc{SKG2Data-Holdout}}

\begin{document}

\title{Spatial Knowledge Graph-Guided Multimodal Synthesis}

\author{
 Yida Xue$^{\textsuperscript{\orcidicon{0000-0002-1970-0678}}}$, Zhen Bi, Jinnan Yang, Jungang Lou, Kehai Chen,~\IEEEmembership{Member,~IEEE},\\ Min Zhang, Huajun Chen$^{\textsuperscript{\orcidicon{0000-0001-5496-7442}}}$, 
 Ningyu Zhang\dag$^{\textsuperscript{\orcidicon{0000-0002-1970-0678}}}$,~\IEEEmembership{Member,~IEEE}

\IEEEcompsocitemizethanks{
\IEEEcompsocthanksitem{
Yida Xue, Zhen Bi, Huajun Chen, Ningyu Zhang are with Zhejiang University, Hangzhou, China.
Ningyu Zhang is the corresponding author. 
\protect (E-mail: \{xueyida, zhangningyu\}@zju.edu.cn.)}
\IEEEcompsocthanksitem{Jinnan Yang is with Nanjing University of Science and Technology, Nanjing, China.
}
\IEEEcompsocthanksitem{Jungang Lou is with Huzhou University, Huzhou, China.
}
\IEEEcompsocthanksitem{Kehai Chen, Min Zhang are with Harbin Institute of Technology, Shenzhen, China.
}
}
}


\maketitle

\begin{abstract}
Recent advances in Multimodal Large Language Models (MLLMs) have significantly enhanced their capabilities; however, their spatial perception abilities remain a notable limitation. To address this challenge, multimodal data synthesis offers a promising solution. Yet, ensuring that synthesized data adhere to spatial common sense is a non-trivial task. Our approach addresses this critical gap by providing a systematic framework for generating spatially coherent data. In this work, we introduce \textbf{{\ourframework}}, a novel multimodal synthesis approach guided by spatial knowledge graphs, grounded in the concept of \textbf{knowledge-to-data} generation. {\ourframework} employs an automated pipeline for constructing \textbf{Spatial Knowledge Graph} (SKG) that effectively captures human-like spatial cognition, including directional and distance relationships. These structured representations then serve as precise guidance for our integrated synthesis pipeline, where a diffusion model generates spatially-consistent images while a MLLM produces corresponding textual descriptions. The automated construction of SKG enables scalable generation of diverse yet realistic spatial configurations, overcoming the limitations of manual data collection and annotation. Extensive experiments demonstrate that data synthesized from diverse types of spatial knowledge, including direction and distance, enhance the spatial perception and reasoning abilities of MLLMs markedly, albeit with a slight cost to their general capabilities. We hope that the idea of knowledge-based data synthesis can advance the development of spatial intelligence\footnote{Code is available at \url{https://github.com/zjunlp/Knowledge2Data}.}.
\end{abstract}


\section{Introduction}
\IEEEPARstart{D}{espite} significant advancements in Multimodal Large Language Models (MLLMs) \cite{zhao2023survey,yin2023survey} for visual processing tasks, a critical limitation persists in their ability to comprehend spatial relationships\cite{rephrase,mmbench,embspatial,scaffolding,eyes_wide,empirical_analysis,Reefknot}. Recent studies demonstrate that even state-of-the-art MLLMs operate with ``eyes wide shut'' \cite{eyes_wide}, exhibiting fundamental deficiencies in visual-spatial comprehension. These models show marked inabilities to accurately estimate relative positions between objects, consistently fail in judging spatial distances, and possess a limited capacity for reasoning about the spatial relationships among multiple objects.
This limitation underscores a significant gap between MLLMs and human capabilities, as humans inherently excel in tasks requiring spatial intelligence.

\begin{figure}[t]
\centering
\includegraphics[width=0.45\textwidth]{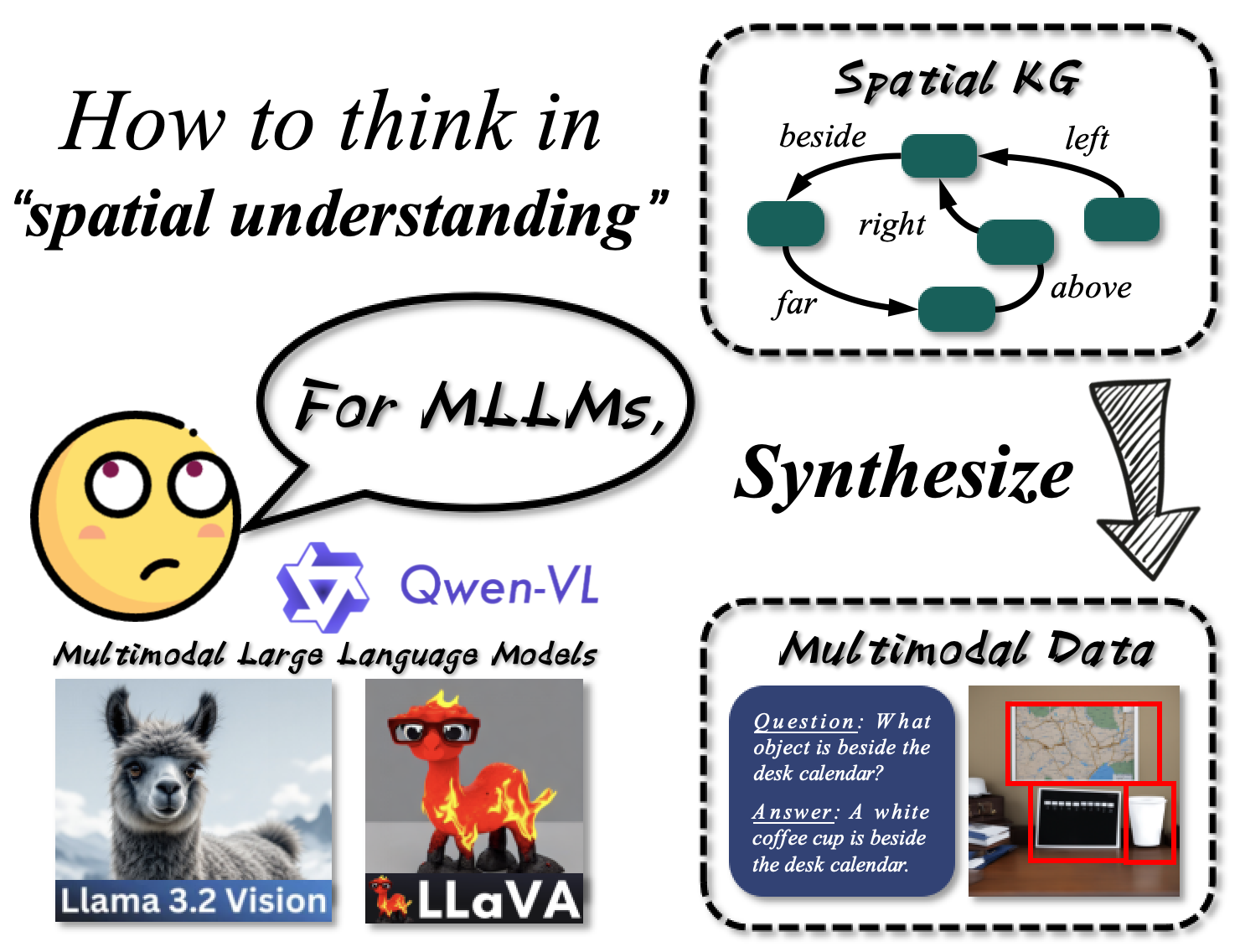}
\caption{
The Spatial Knowledge Graph (SKG) serves as a structured representation of object attributes and spatial relationships. 
Guided by the SKG, {\ourframework} generates images and multimodal data, grounded in the concept of knowledge-to-data generation.
}
\label{fig:intro}
\end{figure}

To address this challenge, multimodal data synthesis offers a promising solution, requiring the generation of large-scale, spatially coherent multimodal data that accurately adheres to real-world constraints. 
A straightforward approach involves developing computational models (or world models) capable of generating continuous synthetic data \cite{embspatial, SAT, reasoningspacemultimodal}. 
However, this method proves inefficient due to redundant data across scenarios and the need for frequent model updates when environments change, resulting in substantial high computational costs for environment updates.
In contrast, human spatial intelligence stems from the acquisition of spatial knowledge by the brain, which allows robust generalization through the application of cognitive frameworks \cite{hodges1999and, herweg2018spatial, peer2021structuring}. 
This insight naturally leads to a new paradigm for multimodal data synthesis: \textbf{modeling spatial knowledge to guide the targeted generation of informative data}. Inspired by this insight, we introduce {\ourframework}, an innovative framework specifically designed to generate multimodal synthetic data for spatial understanding as illustrated in Figure~\ref{fig:intro}. The core idea is to leverage a specialized knowledge graph as a structured spatial representation to guide multimodal data generation. Specifically, we construct a Spatial KG (SKG) by employing GPT-4o to characterize objects and model their spatial relationships, thereby establishing a structured representation of spatial knowledge. The resulting SKG, which encodes object attributes and their spatial relationships, then serves as the key control mechanism in {\ourframework} to direct the generation of corresponding images and textual data.

We automatically synthesize SKG-based spatial multimodal data and design three ways to ensure the quality of synthetic data: human verification, external knowledge base enhancement, and automatic multimodal agent verification. After filtering, we construct a multimodal instruction dataset, which contains 974 nodes and 4,731 triplets encoding spatial relationships.
To ensure a fair comparison of results and prevent data contamination, we also use {\ourframework} to synthesize an evaluation dataset, referred to as {\testdata}, which includes 120 images and 566 single-choice questions. We fine-tune the LLaVA\cite{liu2024llavanext} and Llama-3.2-Vision\cite{llama3.2} models using the synthesized data. The fine-tuned models show performance improvements on {\testdata} and other spatial understanding benchmarks, while maintaining their general abilities. Additionally, we carry out comprehensive experiments to examine the effect of different spatial relation types, dataset sizes, and object densities on model performance.
Our work makes the following principal contributions:

\begin{itemize}
\item We propose {\ourframework}, a novel framework for generating multimodal synthetic data for spatial understanding tasks. It uses a SKG to guide the creation of high-quality multimodal data that reflects real-world spatial constraints, addressing the critical limitation of existing MLLMs in comprehending spatial relationships.
\item We automatically construct a multimodal instruction dataset and an evaluation dataset {\testdata} based on the {\ourframework}. These datasets are rich and diverse, providing comprehensive training data and a fair benchmark for assessing the spatial understanding capabilities of MLLMs.
\item We demonstrate the effectiveness of our synthesized data through extensive experiments. Fine-tuning MLLMs with our data significantly improves their performance on spatial understanding tasks while only slightly reducing their general capabilities. Our findings also reveal that directional knowledge is crucial for enhancing spatial perception and that increasing object density in the data improves fine-grained recognition.
\end{itemize}

\section{Related Work}
\subsection{Multimodal Large Language Models} 
The remarkable success of Large Language Models (LLMs)~\cite{gpt3,gpt4,llama2023} has driven significant advances in Multimodal Large Language Models (MLLMs)~\cite{mPLUG,liu2024llavanext,MiniGPT-v2,llama3.2,gemini,qwenvl,gpt4v,gpt4o}. These MLLMs typically employ transformer-based architectures with specialized visual encoders, such as ViT~\cite{vit} or CLIP~\cite{clip}, that project visual inputs into the language model's embedding space. This architectural innovation enables models to process and understand both text and visual information through a unified framework, achieving state-of-the-art performance on visual tasks.

The development of comprehensive evaluation benchmarks has been crucial for measuring progress in this field. General-purpose multimodal benchmarks like MMVet~\cite{MMVet} and MMStar~\cite{mmstar} assess broad capabilities including visual recognition, text generation, and commonsense reasoning. More specialized benchmarks like HallusionBench~\cite{Hallusionbench} and POPE~\cite{POPE} focus on identifying and quantifying hallucinations in MLLMs.
Recent work has highlighted spatial reasoning as a particularly challenging capability for MLLMs. This has led to the creation of dedicated spatial understanding benchmarks~\cite{whatsup,embspatial,eyes_wide} that systematically evaluate abilities like relative position estimation, distance judgment, and multi-object spatial relationships. These benchmarks reveal that while MLLMs excel at many visual-language tasks, their spatial comprehension still lags significantly behind human performance, particularly in complex real-world scenarios requiring spatial reasoning.

\subsection{Synthetic Data Generation}
Synthetic data has garnered significant attention as a novel solution to the challenges of obtaining large, diverse, and high-quality datasets~\cite{sd_survey0,sd_survey1,sd_survey2,datagenerators}, particularly in domains where real-world data collection is expensive, privacy-sensitive, or inherently limited. Multimodal data synthesis represents an advanced and rapidly evolving field with transformative potential across computer vision, natural language processing, and human-computer interaction applications. This paradigm involves the systematic creation of integrated multimodal data, combining text, images, audio, and sensor modalities through algorithmic generation processes. Specifically, for current MLLMs, many works focus on synthesizing data in both image and text modalities to train or test models.

Many existing studies primarily rely on pre-existing images~\cite{
smqg, fg_network, MathLLaVA}, occasionally incorporating additional image information such as captions, bounding boxes, or OCR input~\cite{liu2023improvedllava,SVIT,LLaVAR}.
However, another area of research transcends generating textual instruction data and focuses on creating high-quality images through various synthesis techniques rather than relying solely on existing image datasets. For example, REACHQA~\cite{reachqa} and Multimodal Self-Struct~\cite{mminstruct} use code to precisely synthesize chart images. Additionally, Scenethesis~\cite{Scenethesis} and EmbSpatial~\cite{embspatial} utilize indoor 3D simulators to generate images, while SynCLR~\cite{synclr} and VisMin~\cite{VisMin} employ diffusion models to produce new images. Furthermore, since LLMs tend to exhibit biases and follow a long-tailed distribution due to their lack of specialization in specific domains~\cite{biases1,biases2}, there is a growing body of research~\cite{FactKB,Knowledge-Infused,InstructProtein,scp,ki} that employs knowledge enhancement techniques, such as KG, to generate higher quality data.

In this work, we propose a method for multimodal data synthesis based on diffusion models and knowledge enhancement techniques. End-to-end text-to-image generative models, such as Stable Diffusion~\cite{stable_diffusion} and Dall-E~\cite{dalle}, often encounter challenges in accurately controlling object positions and quantities. In contrast, layout-based diffusion models, such as GLIGEN-based models~\cite{lmd,GLIGEN}, achieve accurate control over object placement and quantities by incorporating bounding boxes or explicit layout guidance. LLM-grounded Diffusion (LMD)~\cite{lmd} utilizes LLMs to generate bounding boxes as layout information for these layout-based diffusion models.
Specifically, our method leverages spatial knowledge graphs to guide the generation of both images and corresponding textual descriptions, ensuring that the synthesized data adheres to real-world spatial constraints. Recent advances in diffusion models enable high-fidelity image generation, while knowledge graphs provide structured representations that maintain semantic and spatial coherence across modalities.



\begin{figure*}[t]
\centering 
\includegraphics[width=0.95\textwidth]{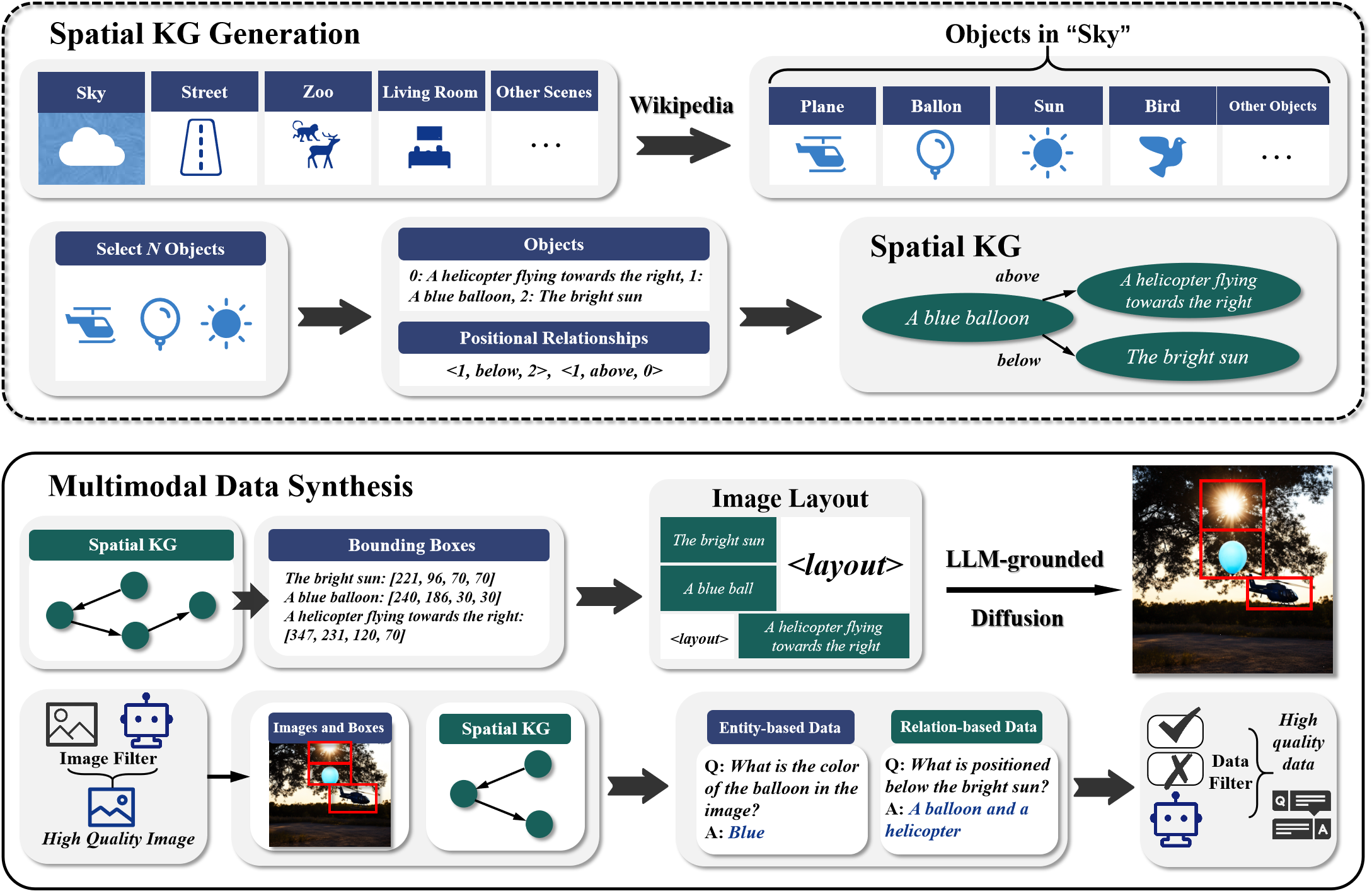}
\caption{
A comprehensive overview of the our framework. Our framework consists of two core modules: \textbf{Spatial KG Generation} and \textbf{Multimodal Data Synthesis}. The \textbf{Spatial KG Generation} module generates an intermediate representation, the Spatial KG, which guides the synthesis of multimodal data. The \textbf{Multimodal Data Synthesis} module is tasked with generating image data and their corresponding textual data.
}
\label{fig:main}
\end{figure*}


\section{Preliminaries}
\subsection{Spatial Knowledge Graph}
In recent years, KGs have been widely applied across various domains, such as FACTKG~\cite{FactKG} for fact verification. 
Unlike traditional KGs, we introduce the Spatial KG (SKG), a specialized KG designed to encode spatial knowledge and guide multimodal data synthesis for spatial tasks. 
The SKG is defined as a KG where nodes represent objects and edges represent spatial relationships through triplets. Formally, SKG can be represented as:

\begin{equation}
G = (E, T)
\end{equation}

where \(E\) denotes the set of nodes (objects) and \(T\) denotes the set of triplets describing spatial relations.
These relationships may include directional relationships (e.g., ``to the left of'' and ``to the right of''), distance relationships (e.g., ``close to'' and ``far away''), or a combination of both. 
SKG represents a collection of objects and their spatial relationships within a specific scene.

In contrast to Scene Graphs~\cite{vg, scene_graph}, which provide a general representation of scenes, the triplets in an SKG are specifically designed to emphasize spatial relations, with nodes \(E\) containing detailed descriptions of objects.
This structured representation facilitates the generation of complex, detailed multimodal data that accurately capture spatial relationships. By systematically modeling spatial relationships through SKG, our method bridges the gap between multimodal synthetic data and human-like spatial understanding.

\subsection{Multimodal Synthetic Data}
The task of multimodal data synthesis involves generating high-quality input-output pairs related to images using a model \(f\). 
For a given image \(i\) (either pre-existing or synthesized), the task uses \(i\), its description \(d\), or a combination of both as inputs to \(f\). 
The model \(f\) then synthesizes a set of input \(Q\) and their corresponding output \(A\).
We focus on generating multimodal data, with an emphasis on capturing spatial relationships between objects. 

Our methodology involves the simultaneous creation of images and their corresponding question-answer pairs through a three-component framework: \textbf{a multimodal generative model \(f_{1}\), a text-to-image generative model \(f_{2}\), and a SKG \(G\)}. 
Inspired by LLM-grounded Diffusion (LMD)~\cite{lmd}, we use \(f_{1}\) to generate bounding boxes \(B\) and description \(d\) under the guidance of \(G\).
These bounding boxes are then integrated into the layout-diffusion model \(f_{2}\) to synthesize images. After image synthesis, \(G\) and \(B\) further assist \(f_{1}\) in generating multiple question-answer pairs \((Q, A)\) relevant to the synthesized image \(i\). This process can be mathematically expressed as follows:
\begin{equation}
\begin{gathered}
(B, d) = f_{1}(G) \\
i = f_{2}(B, d, G) \\
(Q, A) = f_{1}(B, G, i)
\end{gathered}
\end{equation}

For implementation, we use GPT-4o and GLIGEN~\cite{GLIGEN} version of LMD as \(f_{1}\) and \(f_{2}\), respectively.

\section{Spatial Knowledge Graph-Guided Data Synthesis}
In this section, we introduce {\ourframework}, a novel approach that takes advantage of a SKG that incorporates spatial positional relationships to guide the synthesis of multimodal data. As shown in Figure \ref{fig:main}, {\ourframework} consists of two modules: \textbf{Spatial KG Generation} and \textbf{Multimodal Data Synthesis}.

\subsection{Spatial Knowledge Graph Generation}
\paragraph{Scenes and Objects Generation}
The real world encompasses a wide variety of scenes, each containing distinct sets of objects. 
To simulate the distribution of scenes and objects in real-world images, we leverage GPT-4o to generate a series of scenes and the objects likely to appear within them. 
Starting with a few scenes as few-shot examples, we use GPT-4o to generate an additional scenes. After human inspection and filtering, we finally obtain 160 safe, diverse, and realistic scenes, significantly enhancing the diversity of our dataset.
For each scene, GPT-4o is further employed to produce a list of plausible objects that could exist in the given context. To improve object generation, particularly for uncommon scenes, we incorporate Wikipedia documents as external knowledge to ensure the trustworthiness and rationality of objects in each scene. This approach ensures that the synthesized objects exhibit distributions closely aligned with real-world images, effectively preventing the generation of incongruous or improbable objects within the specified scenes. Through observation of the generated objects, we find that most of the objects appear in high frequency in each real scene.
Formally, we first employ \(f_1\) to generate \(m\) general scenes, denoted as \( \{s_1, s_2, \dots, s_m\} \). For each scene \(s_i\), we then generate \(n\) objects that are likely to appear in that scene, expressed as:
\begin{equation}
\{o_{i1}, o_{i2}, \dots, o_{in}\} = f_1(s_i)
\end{equation}

\paragraph{Spatial Knowledge Graph Construction}
Due to the diverse rules and knowledge governing spatial relationships between objects in the real world, we employ GPT-4o to generate positional relationships that align with real-world spatial cognition. Additionally, we introduce the SKG as an intermediary representation to facilitate the generation of high-quality multimodal data for spatial understanding tasks. The SKG is constructed using GPT-4o's capabilities to synthesize entities and their spatial relationship triplets.

In the initial phase, \(f_1\) is used to select a subset \(\{o_{ij_1}, o_{ij_2}, \ldots, o_{ij_k}\}\) of \(k\) candidate objects from the original set \(\{o_{i1}, o_{i2}, \ldots, o_{in}\}\). These selected objects are utilized to construct a specific SKG. The subset \(C_k\) is formally represented as:

\begin{equation}
\begin{gathered}
C_k = \{o_{ij_1}, o_{ij_2}, \ldots, o_{ij_k}\}, \\
C_k \subseteq \{o_{i1}, o_{i2}, \ldots, o_{in}\}, \\
|C_k| = k
\end{gathered}
\end{equation}

\begin{figure*}
\centering 
\includegraphics[width=0.95\textwidth]{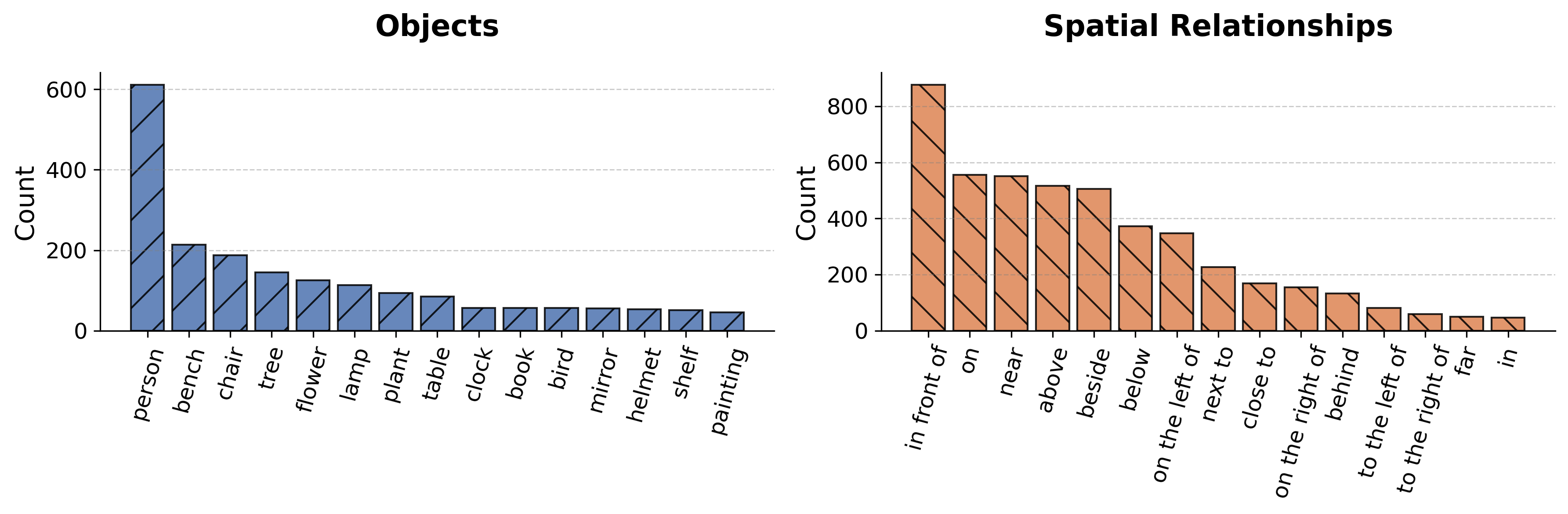}
\caption{
Distribution of top 15 objects and spatial relationships. There are a total of 974 objects and 95 relationships.
}
\label{fig:combined}
\end{figure*}

\begin{figure*}[htb]
\centering 
\includegraphics[width=1\textwidth]{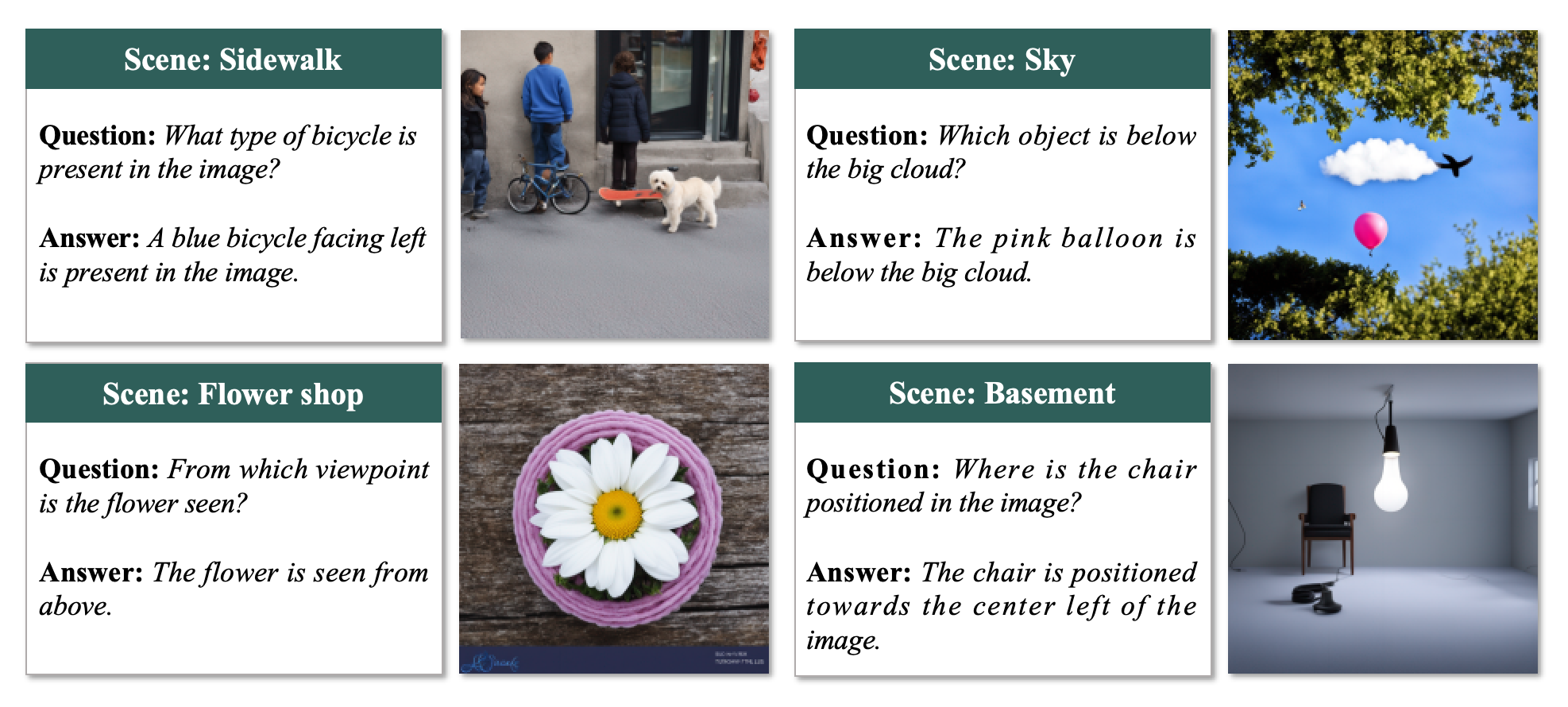}
\caption{
Training data cases featuring diverse spatial relationships, synthesized through our proposed method.
}
\label{fig:train_case}
\end{figure*}

Next, each object in \(C_k\) undergoes an attribute enrichment process via \(f_1\), which adds quantities and attributes such as color, orientation, and material. This results in a set of \(l\) entities, denoted as \(E_l\), with rich descriptive attributes. In \(E_l\), each entity represents a single object with a quantity of one and clearly defined attributes. If objects within the same category have a quantity greater than one, they are divided into multiple entities. For example, a selected object ``Balloon'' with a higher quantity could be expanded into two distinct entities which reflect the addition of specific attributes by GPT-4o: ``A blue balloon'' and ``A yellow balloon''. This approach ensures precise control over the number of entities, maintaining consistency with the objects in the subsequently generated images.

\begin{equation}
\begin{gathered}
E_l = \{e_{ij_1}, e_{ij_2}, \ldots, e_{ij_l}\} = f_1(C_k), \\
|E_l| = l \geq k
\end{gathered}
\end{equation}

The entities in \(E_l\) serve as the nodes of the SKG \(G\). The subsequent task involves synthesizing spatial relationships between pairs of entities in \(E_l\). This is achieved using \(f_1\), which generates a set of relationship triplets \(T_l \):

\begin{equation}
\begin{gathered}
T_l = \{(e_{ij_a}, r, e_{ij_b}) \mid e_{ij_a}, e_{ij_b} \in E_l, \\
a \neq b, r = f_1(e_{ij_a}, e_{ij_b})\}
\end{gathered}
\end{equation}

Each triplet \((e_{ij_a}, r, e_{ij_b})\) represents a unique pair of entities \(e_{ij_a}\) and \(e_{ij_b}\), along with their spatial relationships \(r\) which could involve directional relationships, distance relationships, or a combination of both. The collection of these triplets \(T_l\) comprehensively captures all pairwise spatial interactions within \(E_l\). By combining the nodes \(E_l\) and the triplet relationships \(T_l\), we construct the \(G\).

\subsection{Multimodal Data Synthesis}
\paragraph{Image Data Generation}
We employ GPT-4o to generate bounding boxes and captions under the guidance of SKG, and seamlessly incorporate these outputs into a GLIGEN-based model.
Recognizing that diffusion models cannot guarantee perfection in every generated image, we generate multiple images for each SKG by varying the random seed.

Specifically, we input the intermediate representations \(E_l\) and \(T_l\) derived from \(G\) directly into \(f\) as guidance. Additionally, we leverage few-shot examples to instruct \(f_1\) on how to generate appropriate bounding boxes \(B_l\).

\begin{equation}
\begin{gathered}
B_l = \{b_{ij_1}, b_{ij_2}, \ldots, b_{ij_l}\} = f_1(E_l, T_l), \\
\end{gathered}
\end{equation}

In order to synthesize data on whether an object exists in the image for subsequent tasks, we randomly select an object from other objects in current scene as a non-existent object \(o_{neg}\), ensuring that this object is excluded from the synthesized image. The elements \(E_l\), \(T_l\), and \(s_i\), along with the optional non-existent object \(o_{neg}\), are input to \(f_1\) to produce the caption \(d\). The formula is as follows:

\begin{equation}
\begin{gathered}
o_{neg} \in \{o \mid o \notin C_k \} \cup \{\emptyset\}, \\
d = f_1(E_l, T_l, s_i, o_{neg})
\end{gathered}
\end{equation}

Finally, we input \(B_l\) and \(d\) into \(f_2\) to generate the final image \(i\).

\begin{equation}
\begin{gathered}
i = f_2(d, B_k)
\end{gathered}
\end{equation}
\paragraph{Image Data Filtering}
During the image generation stage, a significant challenge is the potential for discrepancies, such as hallucinations, which can occur during the transformation of the SKG into bounding boxes and captions. Additionally, the image generation model may introduce its own inaccuracies. These issues can result in inconsistencies between the generated image and the original SKG. To address this challenge, we conduct an image validation process using GPT-4o as a image filtering agent. This process is aimed at evaluating and ensuring that the generated image aligns with the SKG. Formally, we utilize \(f_1\) to assess the alignment of an image \(i\) with the expected entities \(E_l\) and relationships \(T_l\). 
If \(f_1\) returns True, the image \(i\) is preserved; otherwise, it is discarded. This process is represented by the following equation:

\begin{equation}
f_1(i, E_l, T_l) =
\begin{cases} 
    \text{True},  & \text{if } i \text{ aligns with} \\ 
                 & \text{both } E_l \text{ and } T_l, \\ 
    \text{False}, & \text{otherwise.}
\end{cases}
\end{equation}

By verifying the attributes and positional relationships of objects in the image through the agent, filter out images that do not match the information in SKG. After filtering, the dataset contains a total of 974 objects and 95 spatial relationships. The distribution of the top 15 objects and spatial relationships in the filtered images is illustrated in Figure \ref{fig:combined}.
\begin{table*}[t]
\centering

\caption{\textbf{Comparison of different models on five benchmarks.}}

\setlength{\tabcolsep}{2.0mm}
\begin{adjustbox}{width=1.0\textwidth,center}
\begin{tabular}{l|c|cc|cc}
\toprule
\textbf{Dataset} & \textbf{LLaVA-1.5} & \textbf{LLaVA-1.6} & \textbf{LLaVA-1.6 (w/ \ourframework)} & \textbf{Llama-3.2-vision} & \textbf{Llama-3.2-vision (w/ \ourframework)} \\
\midrule
KG2Data-Holdout & 62.2 & 68.6 & 70.1 \textcolor{darkred}{(+1.5)} & 73.3 & 74.7 \textcolor{darkred}{(+1.4)} \\
COCO-Spatial & 52.4 & 75.4 & 79.3 \textcolor{darkred}{(+3.9)} & 45.9 & 59.8 \textcolor{darkred}{(+13.9)} \\
MMVP & 24.7 & 32.0 & 36.7 \textcolor{darkred}{(+4.7)} & 29.3 & 30.7 \textcolor{darkred}{(+1.4)} \\
\midrule
MMStar & 33.1 & 37.6 & 36.7 \textcolor{darkgreen}{(-0.9)} & 49.8 & 48.1 \textcolor{darkgreen}{(-1.7)} \\
HallusionBench & 27.6 & 27.6 & 27.2 \textcolor{darkgreen}{(-0.4)} & 40.3 & 45.1 \textcolor{darkred}{(+4.8)} \\
\midrule
Average & 40.0 & 48.2 & 50.0 \textcolor{darkred}{(+1.8)} & 47.7 & 51.7 \textcolor{darkred}{(+4.0)} \\
\bottomrule
\end{tabular}
\end{adjustbox}
\label{tab:main-results}  
\end{table*}

Examples of the training data generated are illustrated in Figure \ref{fig:train_case}.

\paragraph{Textual Data Generation}
The training data for textual data is primarily composed of open-ended question-answer pairs, while the evaluation data predominantly consists of single-choice questions. To enhance GPT-4o's understanding of object attributes and spatial relationships in images, we leverage the SKG to guide the generation of question-answer pairs. This approach not only improves the accuracy of the data generation process but also reduces the occurrence of hallucinations. Furthermore, it enables the creation of more diverse and complex questions tailored for spatial understanding tasks. These questions are categorized into two types based on their primary focus: \textbf{Entity-Based Data} and \textbf{Relation-Based Data}.

\textbf{Entity-Based Data} focuses on the existence, attributes, and quantities of objects in a given image. The construction of Entity-Based Data is primarily guided by the entity set \(E_l\), along with explicitly defined non-existent objects \(o_{neg}\). This type of data serves as general visual question-answering data, maintaining the general capabilities of MLLMs without requiring additional external datasets. In contrast, \textbf{Relation-Based Data} emphasizes the spatial relationships between objects. The generation of Relation-Based Data is based on the spatial relationship triplets \(T_l\) and generated bounding boxes \(B_l\). Relation-Based Data is specifically designed to improve the spatial reasoning capabilities of MLLMs.
The generation process for each question \(x\) and its corresponding answer \(y\) is formalized as follows:
\begin{equation}
\begin{gathered}
x = f_1(i, B_l, E_l, T_l), \\
y = f_2(i, B_l, E_l, T_l, x)
\end{gathered}
\end{equation}

\paragraph{Textual Data Filtering}
For each generated textual data, we employ GPT-4o as a text filtering agent to verify its accuracy. To avoid excessive reliance on the information provided in the SKG, we withhold SKG during the textual data verification phase. Instead, GPT-4o relies solely on the image to verify the correctness of the provided reference answers. The evaluation is performed using \(f_1\), which determines the correctness of the answer \(y\) for the question \(x\) based on the image \(i\). If \(f_1\) returns True, the textual data \((x, y)\) is preserved:

\begin{equation}
f_1(i, x, y) = 
\begin{cases} 
    \text{True} & \text{if } y \text{ is correct}, \\
    \text{False} & \text{otherwise.}
\end{cases}
\end{equation}

After filtering, data pairs that the GPT-4o cannot correctly answer solely by providing images can be filtered.

\begin{figure}
\centering 
\includegraphics[width=0.48\textwidth]{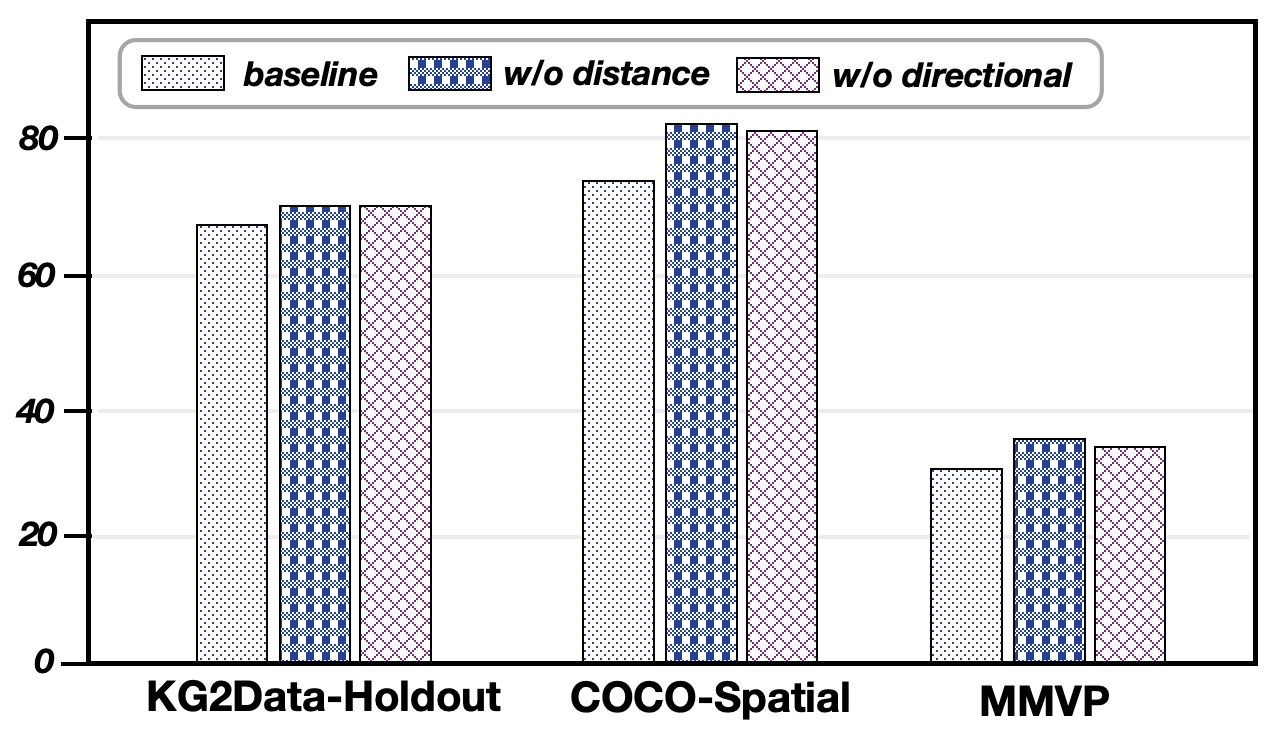}
\caption{
The result of removing specific positional relationship data with the same amount of data.
}
\label{fig:table2}
\end{figure}
\section{Experiments}
\subsection{Experimental Settings}
\paragraph{Baselines}
We conduct experiments on three open-source MLLMs: LLaVA-1.5-7B~\cite{liu2023improvedllava}, LLaVA-1.6-7B~\cite{liu2024llavanext}, and Llama-3.2-Vision-11B~\cite{llama3.2}. Both LLaVA-1.5 and LLaVA-1.6 use the Vicuna-7B~\cite{vicuna2023} LLM as their LLM backbone.

\paragraph{Benchmarks}
For a comprehensive evaluation, we examine all models across two distinct task categories.
The first category encompasses benchmarks that are either directly or partially related to spatial understanding. This includes the publicly available benchmark MMVP~\cite{eyes_wide} and COCO-Spatial~\cite{whatsup}. For COCO-Spatial, we utilize a subset named Two-obj that focuses on the positional relationships of two objects. We then convert the original positive and negative sample pairs into single-choice questions, which are more conducive to MLLMs testing. We leave a portion of the synthesized image to generate test questions called {\testdata}, which are also used for evaluation. The second category involves general capability testing benchmarks for MLLMs, including MMStar~\cite{mmstar} and HallusionBench~\cite{Hallusionbench}. 

\paragraph{Implementation Details}
We fine-tune both LLaVA-v1.6 and Llama-3.2-Vision using training data. These models consist of a vision encoder, a projector, and an LLM backbone. During training, we tune the LLM backbone with LoRA modules~\cite{lora}, keeping the projector and vision encoder parameters fixed. 

The experimental setup employs LoRA for fine-tuning with a global batch size of 64. Training occurs on NVIDIA A800 GPUs using a learning rate of $1 \times 10^{-5}$. The LoRA configuration includes a rank ($r$) of 16, an alpha ($\alpha$) value of 32, and a dropout rate of 0.05. The models undergo training for 2 epochs with mixed precision enabled to optimize computational efficiency. These parameters ensure stable training while maintaining memory efficiency and model performance.

\begin{figure}
\centering 
\includegraphics[width=0.48\textwidth]{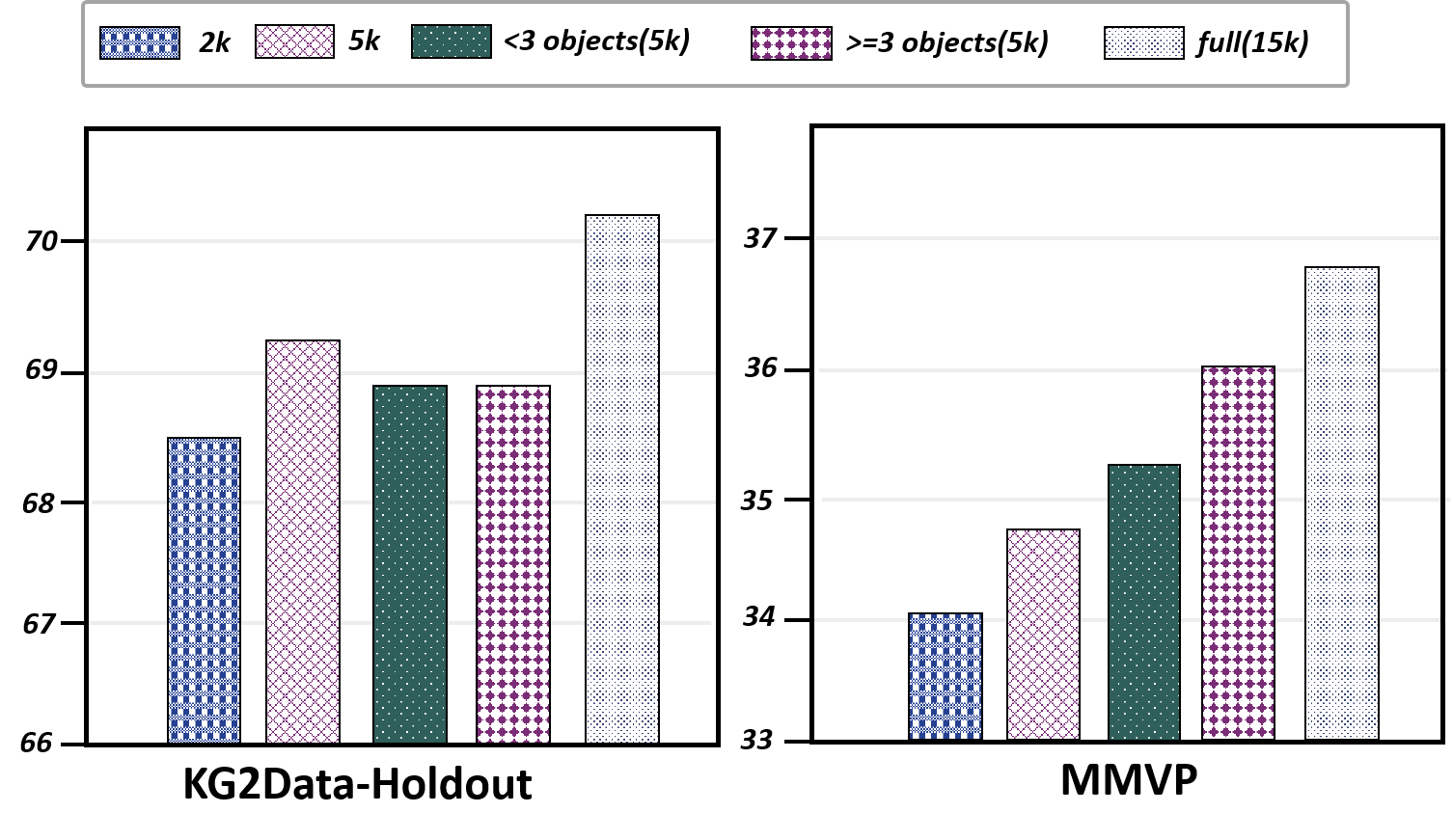}
\caption{
The influence of the number of objects and the quantity of data on experimental results.
}
\label{fig:table3}
\end{figure}

\subsection{Main Results}

Table \ref{tab:main-results} presents a comprehensive comparison of multiple models evaluated across five benchmarks, highlighting the strengths of {\ourframework}. 

\paragraph{Spatial Understanding Results}
Fine-tuned LLaVA-1.6 and fine-tuned Llama-3.2-vision demonstrate marked improvements across several challenging benchmarks for spatial tasks, highlighting the efficacy of our methods. On the KG2Data-Holdout benchmark, Fine-tuned LLaVA-1.6 achieves a score of 70.1\textcolor{darkred}{\textbf{(+1.5)\textuparrow}}, while Fine-tuned Llama-3.2-vision reaches 74.7\textcolor{darkred}{\textbf{(+1.4)\textuparrow}}, indicating robust performance enhancements. For the COCO-Spatial benchmark, Fine-tuned LLaVA-1.6 outperforms with a score of 79.3\textcolor{darkred}{\textbf{(+3.9)\textuparrow}}, and Fine-tuned Llama-3.2-vision impressively achieves 59.8\textcolor{darkred}{\textbf{(+13.9)\textuparrow}}, suggesting significant spatial understanding advancements. In the MMVP benchmark, Fine-tuned LLaVA-1.6 shows notable improvement with a score of 36.7\textcolor{darkred}{\textbf{(+4.7)\textuparrow}}, while Fine-tuned Llama-3.2-vision also makes gains, reaching 30.7\textcolor{darkred}{\textbf{(+1.4)\textuparrow}}. In summary, our experiments show that synthesized training data enhances MLLMs' spatial understanding, even without scene or object generation tailored for these spatial datasets.
\paragraph{General Visual Understanding Results.}
We present the results of our models across several general visual understanding benchmarks: MMStar, MMVet, and HallusionBench.
In the MMStar benchmark, fine-tuned LLaVA-1.6 slightly decreases to 36.7, reflecting a minor decline of \textcolor{darkgreen}{0.9\textdownarrow}, while fine-tuned Llama-3.2-vision shows a slight decrease to 48.1, marking a reduction of \textcolor{darkgreen}{1.7\textdownarrow}. Interestingly, in the HallusionBench benchmark, fine-tuned LLaVA-1.6 experiences a negligible decrease to 27.2, showing a marginal drop of \textcolor{darkgreen}{0.4\textdownarrow}, while fine-tuned Llama-3.2-vision demonstrates a remarkable improvement, advancing to 45.1\textcolor{darkred}{\textbf{(+4.8)\textuparrow}}, indicating enhanced performance in hallucination task. Overall, our results demonstrate that training LLaVA-1.6 and Llama-3.2-vision with synthesized data does not significantly degrade their general visual understanding abilities.
\paragraph{Average Results.}
Many current works have found that training on datasets with specific domains or abilities can affect the performance of models in other domains and general domains, leading to an overall performance declines~\cite{sft1,sft2,sft3}.
For our method, the fine-tuned LLaVA-1.6 model achieves an average score of 50.0 (\textcolor{darkred}{+1.8}), while the fine-tuned Llama-3.2-vision model reaches 51.7 (\textcolor{darkred}{+4.0}). These improvements confirm that {\ourframework} effectively enhances spatial reasoning while maintaining performance in general visual understanding tasks.

\begin{figure*}[htb]
\centering 
\includegraphics[width=0.95\textwidth]{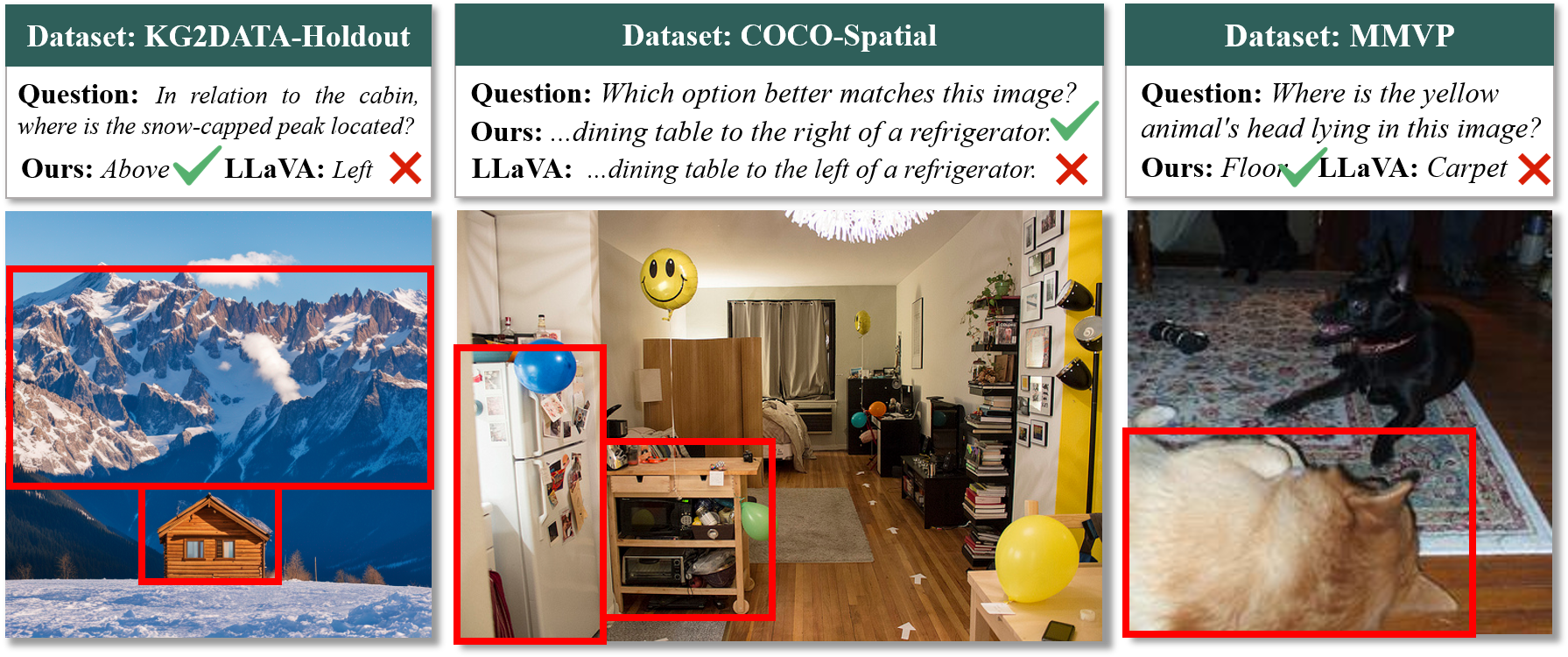}
\caption{
Cases analysis: In spatially related datasets, the trained model enhances spatial understanding abilities.
}
\label{fig:case}
\end{figure*}

\subsection{Ablation and Analysis}
To evaluate the impact of the synthesized data generated by {\ourframework} on model performance, we conducted a comprehensive ablation study using LLaVA-1.6. This study focuses on three key aspects: (1) \textbf{the influence of different relationship types}, (2) \textbf{the effect of varying data quantities}, and (3) \textbf{the impact of the number of objects}. 

For the first aspect,  GPT-4 is utilized to categorize the positional relationships in the training data into two types: distance relationships and directional relationships. The models are then trained separately on each type. To minimize the influence of data quantity on the experimental results, we sample 2,000 data from the training dataset after eliminating one type of positional relationship. The detailed experimental results are depicted in Figure \ref{fig:table2}. For the second aspect, we randomly sample 2k and 5k from the 15k training data. For the third aspect, we regulate the data volume to 5k to examine the influence of the number of objects on the model performance within SKG. Setting the limit to 3 objects, we categorize the numbers into two groups: (1) counts that are greater than or equal to 3 and (2) those that are less than 3. The results of second and third aspects are shown in Figure \ref{fig:table3}. The experimental results shown in Figure \ref{fig:table2} guide us towards \textbf{Finding 1} and \textbf{Finding 2}. Additionally, \textbf{Finding 3} is derived from the experimental results observed in Figure \ref{fig:table3}.

\paragraph{Finding 1. Directional knowledge plays a crucial role in enhancing the MLLM's spatial perception capabilities}
We observe that training data containing directional relationships leads to more significant performance improvements compared to data with distance relationships across three spatially correlated datasets. 
This suggests that MLLMs may be more sensitive to directional knowledge and prioritize spatial perception and understanding through directional relationships. 
A potential explanation is that directional knowledge, compared to distance knowledge, exhibits greater variability, providing stronger spatial signals that are more effective in enhancing model performance.
\paragraph{Finding 2. Data synthesized from two types of spatial knowledge, including direction and distance, exhibit generalization ability}
It can be observed that the use of two types of spatial relationship data can enhance performance across three datasets. Significantly, the COCO-Spatial dataset primarily targets directional understanding, and employing data that solely comprises distance relationships for training models can also enhance performance on this dataset. This improvement could possibly be attributed to the varied spatial correlations present in the data synthesized through the spatial knowledge in SKG, and the generalization of an array of positional relationships. This phenomenon underscores the superiority of knowledge-based data synthesis over previous methods.

\paragraph{Finding 3. Increasing the number of objects improves the performance of MLLMs in visual detail tasks}
Our experiments demonstrate that synthesizing data with varying quantities of objects improves spatial understanding, with greater gains observed as the number of objects increases in MMVP benchmark. Notably, training with more objects significantly enhances fine-grained recognition capabilities, as reflected in improved performance on the detail-oriented MMVP benchmark.
We attribute these improvements to the inherent complexity of multi-object scenes, which expose the model to richer spatial relationships and constraints during learning. By leveraging the structured spatial knowledge embedded in SKG, our proposed {\ourframework} systematically generates multimodal training data with diverse and spatially plausible object arrangements, enabling more effective learning of visual and spatial reasoning.

\paragraph{Noise and Error Analysis}
Despite the implementation of automatic filtering, the synthesized data may still contain noise. This noise arises from potential mismatches between images and text in diffusion models, as well as the inherent possibility of hallucinations in MLLMs.
In addition, due to the inherent limitations of current diffusion models, generated images often show problems like distortion and inconsistencies in object type, attributes, quantity, or spatial arrangement.
Creating realistic images becomes much harder as the SKG becomes more complex. Specifically, when the SKG grows larger—with more nodes representing objects and more edges encoding their relationships—the image generation task becomes significantly more difficult. 
Therefore, the various SKG graph structures currently sampled for synthesizing training data are not too complex.
In some real-world situations where scenes are packed with many objects and complex spatial relationships, our method may have difficulty accurately generating these complex data.

\begin{figure*}[htb]
\centering 
\includegraphics[width=0.95\textwidth]{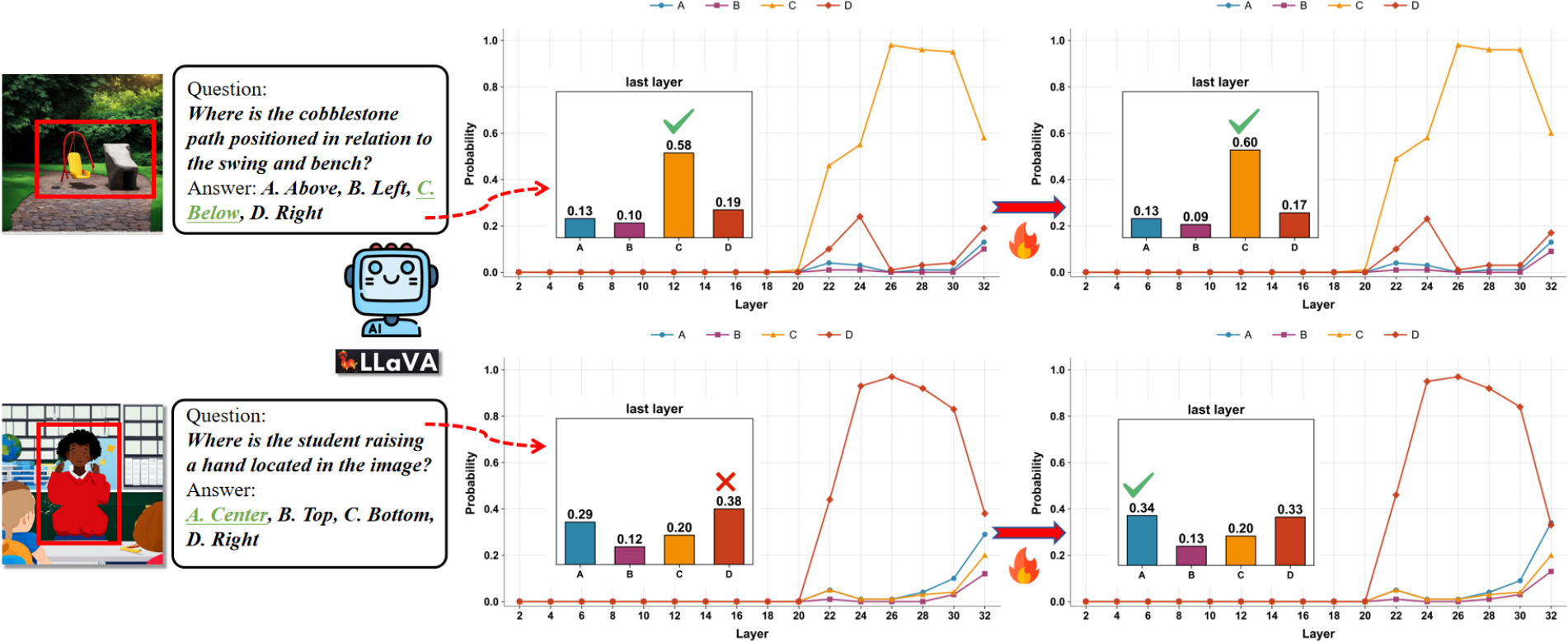}
\caption{
The mechanism of changes in the LLaVA model before and after training. We use two cases to calculate the probability of tokens representing various options at each layer of the transformer, which reveals that the trained model improves the probability of correct options in the last few layers, especially in the last layer.
}
\label{fig:token}
\end{figure*}

\paragraph{Discussion on Knowledge-to-Data Synthesis}
As shown in our experiments, increasing the volume of training data generally leads to improved model performance. Moreover, we observe that high-quality data is important and a small amount of high-quality training data can significantly improve the performance of models in specific fields, which is in line with the opinion of ``less is more''~\cite{limoalignment,limoreasoning}.
As illustrated in the three cases of Figure \ref{fig:case}, the trained model demonstrates significantly improved accuracy in spatially related datasets.
To further investigate how the LLaVA model evolves during training, we select two types of case: one that the model answers correctly even before training and another that it answers correctly only after training.
We analyze internal representations~\cite{representations1,representations2,representations3} in different layers by processing these inputs through the model.
An LM-head attached to each layer produces logits for the first generated token, which are then converted into probabilities via softmax. As shown in Figure \ref{fig:token}, the token probability distribution remains largely unchanged in the previous transformer layers. However, in the last few transformer layers, especially in the last layer, the probability of correct token is rapidly improved.

The core idea of our work is to synthesize data through automated modeling knowledge to ensure the rationality of the data itself. 
The data generated through spatial knowledge is consistent with human spatial cognition, ensuring the relevance and reliability of the data. 
In addition, to synthesize additional data, we can either randomly sample more data from the SKG or expand the scenes to further enrich the SKG. Most existing simulators~\cite{simulator1,simulator2,simulator3,simulator4,simulator5,simulator6,simulator7} are designed for specific environments, and image synthesis via code is largely confined to mathematical visualizations or web-based content. Consequently, such approaches are unable to effectively generate multimodal spatial data beyond these limited contexts. In contrast, our method supports effective data generation across a wide variety of scenarios.

\section{Conclusion}
In this work, we propose {\ourframework}, a novel multimodal synthetic data generation method that leverages Spatial Knowledge Graph to synthesize images and question-answer pairs, addressing the lack of high-quality data for spatial understanding tasks. Using {\ourframework}, we create a multimodal instruction dataset and a benchmark, {\testdata}, to enhance and evaluate the spatial understanding capabilities of MLLMs. Extensive experiments analyze the effectiveness of {\ourframework} and the impact of spatial relationship types, dataset scales, and object densities. Our results show that {\ourframework} significantly improves MLLMs' performance on spatial reasoning tasks.
Furthermore, the automatic construction of Spatial Knowledge Graphs and data synthesis pipeline offers a flexible framework that can be extended to various domains requiring precise spatial representations.

\section*{Acknowledgments}
We would like to express gratitude to the anonymous reviewers for their hard work and kind comments.
This work was supported by the National Natural Science Foundation of China (No. NSFCU23B2055, No. 62576307, No.62506128), the Fundamental Research Funds for the Central Universities (226-2023-00138), 
Zhejiang Provincial Natural Science Foundation of China under Grant (No. LRG25F030003 and LQN25F020023), Yongjiang Talent Introduction Programme (2021A-156-G), Ningbo Natural Science Foundation (2024J020), Information Technology Center and State Key Lab of CAD\&CG, Zhejiang University.

\bibliographystyle{IEEEtran}
\bibliography{custom}

\begin{thebibliography}{10}
\providecommand{\url}[1]{#1}
\csname url@samestyle\endcsname
\providecommand{\newblock}{\relax}
\providecommand{\bibinfo}[2]{#2}
\providecommand{\BIBentrySTDinterwordspacing}{\spaceskip=0pt\relax}
\providecommand{\BIBentryALTinterwordstretchfactor}{4}
\providecommand{\BIBentryALTinterwordspacing}{\spaceskip=\fontdimen2\font plus
\BIBentryALTinterwordstretchfactor\fontdimen3\font minus \fontdimen4\font\relax}
\providecommand{\BIBforeignlanguage}[2]{{%
\expandafter\ifx\csname l@#1\endcsname\relax
\typeout{** WARNING: IEEEtran.bst: No hyphenation pattern has been}%
\typeout{** loaded for the language `#1'. Using the pattern for}%
\typeout{** the default language instead.}%
\else
\language=\csname l@#1\endcsname
\fi
#2}}
\providecommand{\BIBdecl}{\relax}
\BIBdecl

\bibitem{zhao2023survey}
W.~X. Zhao, K.~Zhou, J.~Li, T.~Tang, X.~Wang, Y.~Hou, Y.~Min, B.~Zhang, J.~Zhang, Z.~Dong \emph{et~al.}, ``A survey of large language models,'' \emph{arXiv preprint arXiv:2303.18223}, 2023.

\bibitem{yin2023survey}
S.~Yin, C.~Fu, S.~Zhao, K.~Li, X.~Sun, T.~Xu, and E.~Chen, ``A survey on multimodal large language models,'' \emph{arXiv preprint arXiv:2306.13549}, 2023.

\bibitem{rephrase}
A.~Prasad, E.~Stengel-Eskin, and M.~Bansal, ``Rephrase, augment, reason: Visual grounding of questions for vision-language models,'' \emph{arXiv preprint arXiv:2310.05861}, 2023.

\bibitem{mmbench}
Y.~Liu, H.~Duan, Y.~Zhang, B.~Li, S.~Zhang, W.~Zhao, Y.~Yuan, J.~Wang, C.~He, Z.~Liu \emph{et~al.}, ``Mmbench: Is your multi-modal model an all-around player?'' \emph{arXiv preprint arXiv:2307.06281}, 2023.

\bibitem{embspatial}
\BIBentryALTinterwordspacing
M.~Du, B.~Wu, Z.~Li, X.~Huang, and Z.~Wei, ``Embspatial-bench: Benchmarking spatial understanding for embodied tasks with large vision-language models,'' \emph{CoRR}, vol. abs/2406.05756, 2024. [Online]. Available: \url{https://doi.org/10.48550/arXiv.2406.05756}
\BIBentrySTDinterwordspacing

\bibitem{scaffolding}
X.~Lei, Z.~Yang, X.~Chen, P.~Li, and Y.~Liu, ``Scaffolding coordinates to promote vision-language coordination in large multi-modal models,'' \emph{arXiv preprint arXiv:2402.12058}, 2024.

\bibitem{eyes_wide}
\BIBentryALTinterwordspacing
S.~Tong, Z.~Liu, Y.~Zhai, Y.~Ma, Y.~LeCun, and S.~Xie, ``Eyes wide shut? exploring the visual shortcomings of multimodal llms,'' in \emph{{IEEE/CVF} Conference on Computer Vision and Pattern Recognition, {CVPR} 2024, Seattle, WA, USA, June 16-22, 2024}.\hskip 1em plus 0.5em minus 0.4em\relax {IEEE}, 2024, pp. 9568--9578. [Online]. Available: \url{https://doi.org/10.1109/CVPR52733.2024.00914}
\BIBentrySTDinterwordspacing

\bibitem{empirical_analysis}
\BIBentryALTinterwordspacing
F.~Shiri, X.~Guo, M.~Far, X.~Yu, R.~Haf, and Y.~Li, ``An empirical analysis on spatial reasoning capabilities of large multimodal models,'' in \emph{Proceedings of the 2024 Conference on Empirical Methods in Natural Language Processing, {EMNLP} 2024, Miami, FL, USA, November 12-16, 2024}.\hskip 1em plus 0.5em minus 0.4em\relax Association for Computational Linguistics, 2024, pp. 21\,440--21\,455. [Online]. Available: \url{https://aclanthology.org/2024.emnlp-main.1195}
\BIBentrySTDinterwordspacing

\bibitem{Reefknot}
\BIBentryALTinterwordspacing
K.~Zheng, J.~Chen, Y.~Yan, X.~Zou, and X.~Hu, ``Reefknot: {A} comprehensive benchmark for relation hallucination evaluation, analysis and mitigation in multimodal large language models,'' \emph{CoRR}, vol. abs/2408.09429, 2024. [Online]. Available: \url{https://doi.org/10.48550/arXiv.2408.09429}
\BIBentrySTDinterwordspacing

\bibitem{SAT}
\BIBentryALTinterwordspacing
A.~Ray, J.~Duan, R.~Tan, D.~Bashkirova, R.~Hendrix, K.~Ehsani, A.~Kembhavi, B.~A. Plummer, R.~Krishna, K.~Zeng, and K.~Saenko, ``{SAT:} spatial aptitude training for multimodal language models,'' \emph{CoRR}, vol. abs/2412.07755, 2024. [Online]. Available: \url{https://doi.org/10.48550/arXiv.2412.07755}
\BIBentrySTDinterwordspacing

\bibitem{reasoningspacemultimodal}
\BIBentryALTinterwordspacing
C.~Li, W.~Wu, H.~Zhang, Y.~Xia, S.~Mao, L.~Dong, I.~Vulić, and F.~Wei, ``Imagine while reasoning in space: Multimodal visualization-of-thought,'' 2025. [Online]. Available: \url{https://arxiv.org/abs/2501.07542}
\BIBentrySTDinterwordspacing

\bibitem{hodges1999and}
J.~R. Hodges, J.~Spatt, and K.~Patterson, ``“what” and “how”: evidence for the dissociation of object knowledge and mechanical problem-solving skills in the human brain,'' \emph{Proceedings of the National Academy of Sciences}, vol.~96, no.~16, pp. 9444--9448, 1999.

\bibitem{herweg2018spatial}
N.~A. Herweg and M.~J. Kahana, ``Spatial representations in the human brain,'' \emph{Frontiers in human neuroscience}, vol.~12, p. 297, 2018.

\bibitem{peer2021structuring}
M.~Peer, I.~K. Brunec, N.~S. Newcombe, and R.~A. Epstein, ``Structuring knowledge with cognitive maps and cognitive graphs,'' \emph{Trends in cognitive sciences}, vol.~25, no.~1, pp. 37--54, 2021.

\bibitem{liu2024llavanext}
\BIBentryALTinterwordspacing
H.~Liu, C.~Li, Y.~Li, B.~Li, Y.~Zhang, S.~Shen, and Y.~J. Lee, ``Llava-next: Improved reasoning, ocr, and world knowledge,'' January 2024. [Online]. Available: \url{https://llava-vl.github.io/blog/2024-01-30-llava-next/}
\BIBentrySTDinterwordspacing

\bibitem{llama3.2}
{Meta AI}, ``Llama 3.2: Revolutionizing edge ai and vision with open, customizable models,'' \url{https://ai.meta.com/blog/llama-3-2-connect-2024-vision-edge-mobile-devices/}, 2024.

\bibitem{gpt3}
T.~B. Brown, B.~Mann, N.~Ryder, M.~Subbiah, J.~Kaplan, P.~Dhariwal, A.~Neelakantan, P.~Shyam, G.~Sastry, A.~Askell \emph{et~al.}, ``Language models are few-shot learners,'' \emph{arXiv preprint arXiv:2005.14165}, 2020.

\bibitem{gpt4}
J.~Achiam, S.~Adler, S.~Agarwal, L.~Ahmad, I.~Akkaya, F.~L. Aleman, D.~Almeida, J.~Altenschmidt, S.~Altman, S.~Anadkat \emph{et~al.}, ``Gpt-4 technical report,'' \emph{arXiv preprint arXiv:2303.08774}, 2023.

\bibitem{llama2023}
\BIBentryALTinterwordspacing
H.~Touvron, T.~Lavril, G.~Izacard, X.~Martinet, M.~Lachaux, T.~Lacroix, B.~Rozi{\`{e}}re, N.~Goyal, E.~Hambro, F.~Azhar, A.~Rodriguez, A.~Joulin, E.~Grave, and G.~Lample, ``Llama: Open and efficient foundation language models,'' \emph{CoRR}, vol. abs/2302.13971, 2023. [Online]. Available: \url{https://doi.org/10.48550/arXiv.2302.13971}
\BIBentrySTDinterwordspacing

\bibitem{mPLUG}
\BIBentryALTinterwordspacing
Q.~Ye, H.~Xu, G.~Xu, J.~Ye, M.~Yan, Y.~Zhou, J.~Wang, A.~Hu, P.~Shi, Y.~Shi, C.~Li, Y.~Xu, H.~Chen, J.~Tian, Q.~Qi, J.~Zhang, and F.~Huang, ``mplug-owl: Modularization empowers large language models with multimodality,'' \emph{CoRR}, vol. abs/2304.14178, 2023. [Online]. Available: \url{https://doi.org/10.48550/arXiv.2304.14178}
\BIBentrySTDinterwordspacing

\bibitem{MiniGPT-v2}
J.~Chen, D.~Zhu, X.~Shen, X.~Li, Z.~Liu, P.~Zhang, R.~Krishnamoorthi, V.~Chandra, Y.~Xiong, and M.~Elhoseiny, ``Minigpt-v2: large language model as a unified interface for vision-language multi-task learning,'' \emph{arXiv preprint arXiv:2310.09478}, 2023.

\bibitem{gemini}
{Gemini Team}, ``Gemini: A family of highly capable multimodal models,'' 2024.

\bibitem{qwenvl}
\BIBentryALTinterwordspacing
J.~Bai, S.~Bai, S.~Yang, S.~Wang, S.~Tan, P.~Wang, J.~Lin, C.~Zhou, and J.~Zhou, ``Qwen-vl: {A} frontier large vision-language model with versatile abilities,'' \emph{CoRR}, vol. abs/2308.12966, 2023. [Online]. Available: \url{https://doi.org/10.48550/arXiv.2308.12966}
\BIBentrySTDinterwordspacing

\bibitem{gpt4v}
\BIBentryALTinterwordspacing
OpenAI, ``{GPT-4} technical report,'' \emph{CoRR}, vol. abs/2303.08774, 2023. [Online]. Available: \url{https://doi.org/10.48550/arXiv.2303.08774}
\BIBentrySTDinterwordspacing

\bibitem{gpt4o}
\BIBentryALTinterwordspacing
------, ``Gpt-4o system card,'' 2024. [Online]. Available: \url{https://arxiv.org/abs/2410.21276}
\BIBentrySTDinterwordspacing

\bibitem{vit}
\BIBentryALTinterwordspacing
A.~Dosovitskiy, L.~Beyer, A.~Kolesnikov, D.~Weissenborn, X.~Zhai, T.~Unterthiner, M.~Dehghani, M.~Minderer, G.~Heigold, S.~Gelly, J.~Uszkoreit, and N.~Houlsby, ``An image is worth 16x16 words: Transformers for image recognition at scale,'' in \emph{9th International Conference on Learning Representations, {ICLR} 2021, Virtual Event, Austria, May 3-7, 2021}, 2021. [Online]. Available: \url{https://openreview.net/forum?id=YicbFdNTTy}
\BIBentrySTDinterwordspacing

\bibitem{clip}
\BIBentryALTinterwordspacing
A.~Radford, J.~W. Kim, C.~Hallacy, A.~Ramesh, G.~Goh, S.~Agarwal, G.~Sastry, A.~Askell, P.~Mishkin, J.~Clark, G.~Krueger, and I.~Sutskever, ``Learning transferable visual models from natural language supervision,'' in \emph{Proceedings of the 38th International Conference on Machine Learning, {ICML} 2021, 18-24 July 2021, Virtual Event}, 2021, pp. 8748--8763. [Online]. Available: \url{http://proceedings.mlr.press/v139/radford21a.html}
\BIBentrySTDinterwordspacing

\bibitem{MMVet}
W.~Yu, Z.~Yang, L.~Li, J.~Wang, K.~Lin, Z.~Liu, X.~Wang, and L.~Wang, ``{MM-Vet}: Evaluating large multimodal models for integrated capabilities,'' \emph{arXiv preprint arXiv:2308.02490}, 2023.

\bibitem{mmstar}
L.~Chen, J.~Li, X.~Dong, P.~Zhang, Y.~Zang, Z.~Chen, H.~Duan, J.~Wang, Y.~Qiao, D.~Lin \emph{et~al.}, ``Are we on the right way for evaluating large vision-language models?'' \emph{arXiv preprint arXiv:2403.20330}, 2024.

\bibitem{Hallusionbench}
\BIBentryALTinterwordspacing
T.~Guan, F.~Liu, X.~Wu, R.~Xian, Z.~Li, X.~Liu, X.~Wang, L.~Chen, F.~Huang, Y.~Yacoob, D.~Manocha, and T.~Zhou, ``Hallusionbench: An advanced diagnostic suite for entangled language hallucination and visual illusion in large vision-language models,'' in \emph{{IEEE/CVF} Conference on Computer Vision and Pattern Recognition, {CVPR} 2024, Seattle, WA, USA, June 16-22, 2024}.\hskip 1em plus 0.5em minus 0.4em\relax {IEEE}, 2024, pp. 14\,375--14\,385. [Online]. Available: \url{https://doi.org/10.1109/CVPR52733.2024.01363}
\BIBentrySTDinterwordspacing

\bibitem{POPE}
\BIBentryALTinterwordspacing
Y.~He, H.~Sun, P.~Ren, J.~Wang, H.~Wang, Q.~Qi, Z.~Zhuang, and J.~Wang, ``Evaluating and mitigating object hallucination in large vision-language models: Can they still see removed objects?'' in \emph{Proceedings of the 2025 Conference of the Nations of the Americas Chapter of the Association for Computational Linguistics: Human Language Technologies, {NAACL} 2025 - Volume 1: Long Papers, Albuquerque, New Mexico, USA, April 29 - May 4, 2025}, 2025, pp. 6841--6858. [Online]. Available: \url{https://aclanthology.org/2025.naacl-long.349/}
\BIBentrySTDinterwordspacing

\bibitem{whatsup}
A.~Kamath, J.~Hessel, and K.-W. Chang, ``What's" up" with vision-language models? investigating their struggle with spatial reasoning,'' \emph{arXiv preprint arXiv:2310.19785}, 2023.

\bibitem{sd_survey0}
\BIBentryALTinterwordspacing
H.~Chang, P.~Chen, T.~Chou, C.~Kao, H.~Yu, Y.~Lin, and Y.~Chen, ``A survey of data synthesis approaches,'' \emph{CoRR}, vol. abs/2407.03672, 2024. [Online]. Available: \url{https://doi.org/10.48550/arXiv.2407.03672}
\BIBentrySTDinterwordspacing

\bibitem{sd_survey1}
\BIBentryALTinterwordspacing
R.~Liu, J.~Wei, F.~Liu, C.~Si, Y.~Zhang, J.~Rao, S.~Zheng, D.~Peng, D.~Yang, D.~Zhou, and A.~M. Dai, ``Best practices and lessons learned on synthetic data for language models,'' \emph{CoRR}, vol. abs/2404.07503, 2024. [Online]. Available: \url{https://doi.org/10.48550/arXiv.2404.07503}
\BIBentrySTDinterwordspacing

\bibitem{sd_survey2}
\BIBentryALTinterwordspacing
L.~Long, R.~Wang, R.~Xiao, J.~Zhao, X.~Ding, G.~Chen, and H.~Wang, ``On llms-driven synthetic data generation, curation, and evaluation: {A} survey,'' in \emph{Findings of the Association for Computational Linguistics, {ACL} 2024, Bangkok, Thailand and virtual meeting, August 11-16, 2024}, L.~Ku, A.~Martins, and V.~Srikumar, Eds.\hskip 1em plus 0.5em minus 0.4em\relax Association for Computational Linguistics, 2024, pp. 11\,065--11\,082. [Online]. Available: \url{https://doi.org/10.18653/v1/2024.findings-acl.658}
\BIBentrySTDinterwordspacing

\bibitem{datagenerators}
\BIBentryALTinterwordspacing
S.~Kim, J.~Suk, X.~Yue, V.~Viswanathan, S.~Lee, Y.~Wang, K.~Gashteovski, C.~Lawrence, S.~Welleck, and G.~Neubig, ``Evaluating language models as synthetic data generators,'' \emph{CoRR}, vol. abs/2412.03679, 2024. [Online]. Available: \url{https://doi.org/10.48550/arXiv.2412.03679}
\BIBentrySTDinterwordspacing

\bibitem{smqg}
\BIBentryALTinterwordspacing
I.~Wu, S.~Jayanthi, V.~Viswanathan, S.~Rosenberg, S.~Pakazad, T.~Wu, and G.~Neubig, ``Synthetic multimodal question generation,'' in \emph{Findings of the Association for Computational Linguistics: {EMNLP} 2024, Miami, Florida, USA, November 12-16, 2024}, Y.~Al{-}Onaizan, M.~Bansal, and Y.~Chen, Eds.\hskip 1em plus 0.5em minus 0.4em\relax Association for Computational Linguistics, 2024, pp. 12\,960--12\,993. [Online]. Available: \url{https://aclanthology.org/2024.findings-emnlp.759}
\BIBentrySTDinterwordspacing

\bibitem{fg_network}
L.~Yuan, Y.~Cai, J.~Xu, Q.~Li, and T.~Wang, ``A fine-grained network for joint multimodal entity-relation extraction,'' \emph{IEEE Transactions on Knowledge and Data Engineering}, 2025.

\bibitem{MathLLaVA}
\BIBentryALTinterwordspacing
W.~Shi, Z.~Hu, Y.~Bin, J.~Liu, Y.~Yang, S.-K. Ng, L.~Bing, and R.~K. wei Lee, ``Math-llava: Bootstrapping mathematical reasoning for multimodal large language models,'' in \emph{Conference on Empirical Methods in Natural Language Processing}, 2024. [Online]. Available: \url{https://api.semanticscholar.org/CorpusID:270710901}
\BIBentrySTDinterwordspacing

\bibitem{liu2023improvedllava}
H.~Liu, C.~Li, Y.~Li, and Y.~J. Lee, ``Improved baselines with visual instruction tuning,'' 2023.

\bibitem{SVIT}
\BIBentryALTinterwordspacing
B.~Zhao, B.~Wu, and T.~Huang, ``{SVIT:} scaling up visual instruction tuning,'' \emph{CoRR}, vol. abs/2307.04087, 2023. [Online]. Available: \url{https://doi.org/10.48550/arXiv.2307.04087}
\BIBentrySTDinterwordspacing

\bibitem{LLaVAR}
\BIBentryALTinterwordspacing
Y.~Zhang, R.~Zhang, J.~Gu, Y.~Zhou, N.~Lipka, D.~Yang, and T.~Sun, ``Llavar: Enhanced visual instruction tuning for text-rich image understanding,'' \emph{CoRR}, vol. abs/2306.17107, 2023. [Online]. Available: \url{https://doi.org/10.48550/arXiv.2306.17107}
\BIBentrySTDinterwordspacing

\bibitem{reachqa}
\BIBentryALTinterwordspacing
W.~He, Z.~Xi, W.~Zhao, X.~Fan, Y.~Ding, Z.~Shan, T.~Gui, Q.~Zhang, and X.~Huang, ``Distill visual chart reasoning ability from llms to mllms,'' \emph{CoRR}, vol. abs/2410.18798, 2024. [Online]. Available: \url{https://doi.org/10.48550/arXiv.2410.18798}
\BIBentrySTDinterwordspacing

\bibitem{mminstruct}
\BIBentryALTinterwordspacing
W.~Zhang, Z.~Cheng, Y.~He, M.~Wang, Y.~Shen, Z.~Tan, G.~Hou, M.~He, Y.~Ma, W.~Lu, and Y.~Zhuang, ``Multimodal self-instruct: Synthetic abstract image and visual reasoning instruction using language model,'' in \emph{Proceedings of the 2024 Conference on Empirical Methods in Natural Language Processing, {EMNLP} 2024, Miami, FL, USA, November 12-16, 2024}, Y.~Al{-}Onaizan, M.~Bansal, and Y.~Chen, Eds.\hskip 1em plus 0.5em minus 0.4em\relax Association for Computational Linguistics, 2024, pp. 19\,228--19\,252. [Online]. Available: \url{https://aclanthology.org/2024.emnlp-main.1072}
\BIBentrySTDinterwordspacing

\bibitem{Scenethesis}
\BIBentryALTinterwordspacing
L.~Ling, C.-H. Lin, T.-Y. Lin, Y.~Ding, Y.~Zeng, Y.~Sheng, Y.~Ge, M.-Y. Liu, A.~Bera, and Z.~Li, ``Scenethesis: A language and vision agentic framework for 3d scene generation,'' 2025. [Online]. Available: \url{https://arxiv.org/abs/2505.02836}
\BIBentrySTDinterwordspacing

\bibitem{synclr}
\BIBentryALTinterwordspacing
Y.~Tian, L.~Fan, K.~Chen, D.~Katabi, D.~Krishnan, and P.~Isola, ``Learning vision from models rivals learning vision from data,'' in \emph{{IEEE/CVF} Conference on Computer Vision and Pattern Recognition, {CVPR} 2024, Seattle, WA, USA, June 16-22, 2024}.\hskip 1em plus 0.5em minus 0.4em\relax {IEEE}, 2024, pp. 15\,887--15\,898. [Online]. Available: \url{https://doi.org/10.1109/CVPR52733.2024.01504}
\BIBentrySTDinterwordspacing

\bibitem{VisMin}
\BIBentryALTinterwordspacing
R.~Awal, S.~Ahmadi, L.~Zhang, and A.~Agrawal, ``Vismin: Visual minimal-change understanding,'' \emph{CoRR}, vol. abs/2407.16772, 2024. [Online]. Available: \url{https://doi.org/10.48550/arXiv.2407.16772}
\BIBentrySTDinterwordspacing

\bibitem{biases1}
\BIBentryALTinterwordspacing
R.~Navigli, S.~Conia, and B.~Ross, ``Biases in large language models: Origins, inventory, and discussion,'' \emph{{ACM} J. Data Inf. Qual.}, vol.~15, no.~2, pp. 10:1--10:21, 2023. [Online]. Available: \url{https://doi.org/10.1145/3597307}
\BIBentrySTDinterwordspacing

\bibitem{biases2}
\BIBentryALTinterwordspacing
Y.~Fei, Y.~Hou, Z.~Chen, and A.~Bosselut, ``Mitigating label biases for in-context learning,'' in \emph{Proceedings of the 61st Annual Meeting of the Association for Computational Linguistics (Volume 1: Long Papers), {ACL} 2023, Toronto, Canada, July 9-14, 2023}, A.~Rogers, J.~L. Boyd{-}Graber, and N.~Okazaki, Eds.\hskip 1em plus 0.5em minus 0.4em\relax Association for Computational Linguistics, 2023, pp. 14\,014--14\,031. [Online]. Available: \url{https://doi.org/10.18653/v1/2023.acl-long.783}
\BIBentrySTDinterwordspacing

\bibitem{FactKB}
\BIBentryALTinterwordspacing
S.~Feng, V.~Balachandran, Y.~Bai, and Y.~Tsvetkov, ``Factkb: Generalizable factuality evaluation using language models enhanced with factual knowledge,'' in \emph{Proceedings of the 2023 Conference on Empirical Methods in Natural Language Processing, {EMNLP} 2023, Singapore, December 6-10, 2023}, H.~Bouamor, J.~Pino, and K.~Bali, Eds.\hskip 1em plus 0.5em minus 0.4em\relax Association for Computational Linguistics, 2023, pp. 933--952. [Online]. Available: \url{https://doi.org/10.18653/v1/2023.emnlp-main.59}
\BIBentrySTDinterwordspacing

\bibitem{Knowledge-Infused}
\BIBentryALTinterwordspacing
R.~Xu, H.~Cui, Y.~Yu, X.~Kan, W.~Shi, Y.~Zhuang, M.~D. Wang, W.~Jin, J.~C. Ho, and C.~Yang, ``Knowledge-infused prompting: Assessing and advancing clinical text data generation with large language models,'' in \emph{Findings of the Association for Computational Linguistics, {ACL} 2024, Bangkok, Thailand and virtual meeting, August 11-16, 2024}, L.~Ku, A.~Martins, and V.~Srikumar, Eds.\hskip 1em plus 0.5em minus 0.4em\relax Association for Computational Linguistics, 2024, pp. 15\,496--15\,523. [Online]. Available: \url{https://doi.org/10.18653/v1/2024.findings-acl.916}
\BIBentrySTDinterwordspacing

\bibitem{InstructProtein}
\BIBentryALTinterwordspacing
Z.~Wang, Q.~Zhang, K.~Ding, M.~Qin, X.~Zhuang, X.~Li, and H.~Chen, ``Instructprotein: Aligning human and protein language via knowledge instruction,'' in \emph{Proceedings of the 62nd Annual Meeting of the Association for Computational Linguistics (Volume 1: Long Papers), {ACL} 2024, Bangkok, Thailand, August 11-16, 2024}, L.~Ku, A.~Martins, and V.~Srikumar, Eds.\hskip 1em plus 0.5em minus 0.4em\relax Association for Computational Linguistics, 2024, pp. 1114--1136. [Online]. Available: \url{https://doi.org/10.18653/v1/2024.acl-long.62}
\BIBentrySTDinterwordspacing

\bibitem{scp}
\BIBentryALTinterwordspacing
Z.~Yang, N.~Band, S.~Li, E.~J. Cand{\`{e}}s, and T.~Hashimoto, ``Synthetic continued pretraining,'' in \emph{The Thirteenth International Conference on Learning Representations, {ICLR} 2025, Singapore, April 24-28, 2025}.\hskip 1em plus 0.5em minus 0.4em\relax OpenReview.net, 2025. [Online]. Available: \url{https://openreview.net/forum?id=07yvxWDSla}
\BIBentrySTDinterwordspacing

\bibitem{ki}
\BIBentryALTinterwordspacing
O.~Ovadia, M.~Brief, R.~Lemberg, and E.~Sheetrit, ``Knowledge-instruct: Effective continual pre-training from limited data using instructions,'' 2025. [Online]. Available: \url{https://arxiv.org/abs/2504.05571}
\BIBentrySTDinterwordspacing

\bibitem{stable_diffusion}
\BIBentryALTinterwordspacing
R.~Rombach, A.~Blattmann, D.~Lorenz, P.~Esser, and B.~Ommer, ``High-resolution image synthesis with latent diffusion models,'' \emph{CoRR}, vol. abs/2112.10752, 2021. [Online]. Available: \url{https://arxiv.org/abs/2112.10752}
\BIBentrySTDinterwordspacing

\bibitem{dalle}
J.~Betker, G.~Goh, L.~Jing, T.~Brooks, J.~Wang, L.~Li, L.~Ouyang, J.~Zhuang, J.~Lee, Y.~Guo \emph{et~al.}, ``Improving image generation with better captions,'' \emph{Computer Science. https://cdn. openai. com/papers/dall-e-3. pdf}, 2023.

\bibitem{lmd}
\BIBentryALTinterwordspacing
L.~Lian, B.~Li, A.~Yala, and T.~Darrell, ``Llm-grounded diffusion: Enhancing prompt understanding of text-to-image diffusion models with large language models,'' \emph{Trans. Mach. Learn. Res.}, vol. 2024, 2024. [Online]. Available: \url{https://openreview.net/forum?id=hFALpTb4fR}
\BIBentrySTDinterwordspacing

\bibitem{GLIGEN}
\BIBentryALTinterwordspacing
Y.~Li, H.~Liu, Q.~Wu, F.~Mu, J.~Yang, J.~Gao, C.~Li, and Y.~J. Lee, ``{GLIGEN:} open-set grounded text-to-image generation,'' in \emph{{IEEE/CVF} Conference on Computer Vision and Pattern Recognition, {CVPR} 2023, Vancouver, BC, Canada, June 17-24, 2023}.\hskip 1em plus 0.5em minus 0.4em\relax {IEEE}, 2023, pp. 22\,511--22\,521. [Online]. Available: \url{https://doi.org/10.1109/CVPR52729.2023.02156}
\BIBentrySTDinterwordspacing

\bibitem{FactKG}
\BIBentryALTinterwordspacing
J.~Kim, S.~Park, Y.~Kwon, Y.~Jo, J.~Thorne, and E.~Choi, ``Factkg: Fact verification via reasoning on knowledge graphs,'' in \emph{Proceedings of the 61st Annual Meeting of the Association for Computational Linguistics (Volume 1: Long Papers), {ACL} 2023, Toronto, Canada, July 9-14, 2023}, A.~Rogers, J.~L. Boyd{-}Graber, and N.~Okazaki, Eds.\hskip 1em plus 0.5em minus 0.4em\relax Association for Computational Linguistics, 2023, pp. 16\,190--16\,206. [Online]. Available: \url{https://doi.org/10.18653/v1/2023.acl-long.895}
\BIBentrySTDinterwordspacing

\bibitem{vg}
\BIBentryALTinterwordspacing
R.~K.~Y. Zhu, O.~Groth, J.~Johnson, K.~Hata, J.~Kravitz, S.~Chen, Y.~Kalantidis, L.~Li, D.~A. Shamma, M.~S. Bernstein, and L.~Fei{-}Fei, ``Visual genome: Connecting language and vision using crowdsourced dense image annotations,'' \emph{Int. J. Comput. Vis.}, vol. 123, no.~1, pp. 32--73, 2017. [Online]. Available: \url{https://doi.org/10.1007/s11263-016-0981-7}
\BIBentrySTDinterwordspacing

\bibitem{scene_graph}
\BIBentryALTinterwordspacing
G.~Zhu, L.~Zhang, Y.~Jiang, Y.~Dang, H.~Hou, P.~Shen, M.~Feng, X.~Zhao, Q.~Miao, S.~A.~A. Shah, and M.~Bennamoun, ``Scene graph generation: {A} comprehensive survey,'' \emph{CoRR}, vol. abs/2201.00443, 2022. [Online]. Available: \url{https://arxiv.org/abs/2201.00443}
\BIBentrySTDinterwordspacing

\bibitem{vicuna2023}
\BIBentryALTinterwordspacing
W.-L. Chiang, Z.~Li, Z.~Lin, Y.~Sheng, Z.~Wu, H.~Zhang, L.~Zheng, S.~Zhuang, Y.~Zhuang, J.~E. Gonzalez, I.~Stoica, and E.~P. Xing, ``Vicuna: An open-source chatbot impressing gpt-4 with 90\%* chatgpt quality,'' March 2023. [Online]. Available: \url{https://lmsys.org/blog/2023-03-30-vicuna/}
\BIBentrySTDinterwordspacing

\bibitem{lora}
\BIBentryALTinterwordspacing
E.~J. Hu, Y.~Shen, P.~Wallis, Z.~Allen{-}Zhu, Y.~Li, S.~Wang, and W.~Chen, ``Lora: Low-rank adaptation of large language models,'' \emph{CoRR}, vol. abs/2106.09685, 2021. [Online]. Available: \url{https://arxiv.org/abs/2106.09685}
\BIBentrySTDinterwordspacing

\bibitem{sft1}
\BIBentryALTinterwordspacing
T.~Chu, Y.~Zhai, J.~Yang, S.~Tong, S.~Xie, D.~Schuurmans, Q.~V. Le, S.~Levine, and Y.~Ma, ``{SFT} memorizes, {RL} generalizes: {A} comparative study of foundation model post-training,'' \emph{CoRR}, vol. abs/2501.17161, 2025. [Online]. Available: \url{https://doi.org/10.48550/arXiv.2501.17161}
\BIBentrySTDinterwordspacing

\bibitem{sft2}
\BIBentryALTinterwordspacing
G.~Dong, H.~Yuan, K.~Lu, C.~Li, M.~Xue, D.~Liu, W.~Wang, Z.~Yuan, C.~Zhou, and J.~Zhou, ``How abilities in large language models are affected by supervised fine-tuning data composition,'' 2024. [Online]. Available: \url{https://arxiv.org/abs/2310.05492}
\BIBentrySTDinterwordspacing

\bibitem{sft3}
\BIBentryALTinterwordspacing
Z.~Gekhman, G.~Yona, R.~Aharoni, M.~Eyal, A.~Feder, R.~Reichart, and J.~Herzig, ``Does fine-tuning llms on new knowledge encourage hallucinations?'' in \emph{Proceedings of the 2024 Conference on Empirical Methods in Natural Language Processing, {EMNLP} 2024, Miami, FL, USA, November 12-16, 2024}, Y.~Al{-}Onaizan, M.~Bansal, and Y.~Chen, Eds.\hskip 1em plus 0.5em minus 0.4em\relax Association for Computational Linguistics, 2024. [Online]. Available: \url{https://doi.org/10.18653/v1/2024.emnlp-main.444}
\BIBentrySTDinterwordspacing

\bibitem{limoalignment}
\BIBentryALTinterwordspacing
C.~Zhou, P.~Liu, P.~Xu, S.~Iyer, J.~Sun, Y.~Mao, X.~Ma, A.~Efrat, P.~Yu, L.~Yu, S.~Zhang, G.~Ghosh, M.~Lewis, L.~Zettlemoyer, and O.~Levy, ``{LIMA:} less is more for alignment,'' in \emph{NeurIPS 2023}, A.~Oh, T.~Naumann, A.~Globerson, K.~Saenko, M.~Hardt, and S.~Levine, Eds., 2023. [Online]. Available: \url{http://papers.nips.cc/paper\_files/paper/2023/hash/ac662d74829e4407ce1d126477f4a03a-Abstract-Conference.html}
\BIBentrySTDinterwordspacing

\bibitem{limoreasoning}
\BIBentryALTinterwordspacing
Y.~Ye, Z.~Huang, Y.~Xiao, E.~Chern, S.~Xia, and P.~Liu, ``Limo: Less is more for reasoning,'' 2025. [Online]. Available: \url{https://arxiv.org/abs/2502.03387}
\BIBentrySTDinterwordspacing

\bibitem{representations1}
\BIBentryALTinterwordspacing
M.~Elbayad, J.~Gu, E.~Grave, and M.~Auli, ``Depth-adaptive transformer,'' in \emph{8th International Conference on Learning Representations, {ICLR} 2020, Addis Ababa, Ethiopia, April 26-30, 2020}.\hskip 1em plus 0.5em minus 0.4em\relax OpenReview.net, 2020. [Online]. Available: \url{https://openreview.net/forum?id=SJg7KhVKPH}
\BIBentrySTDinterwordspacing

\bibitem{representations2}
\BIBentryALTinterwordspacing
T.~Schuster, A.~Fisch, J.~Gupta, M.~Dehghani, D.~Bahri, V.~Tran, Y.~Tay, and D.~Metzler, ``Confident adaptive language modeling,'' in \emph{Advances in Neural Information Processing Systems 35: Annual Conference on Neural Information Processing Systems 2022, NeurIPS 2022, New Orleans, LA, USA, November 28 - December 9, 2022}, S.~Koyejo, S.~Mohamed, A.~Agarwal, D.~Belgrave, K.~Cho, and A.~Oh, Eds., 2022. [Online]. Available: \url{http://papers.nips.cc/paper\_files/paper/2022/hash/6fac9e316a4ae75ea244ddcef1982c71-Abstract-Conference.html}
\BIBentrySTDinterwordspacing

\bibitem{representations3}
\BIBentryALTinterwordspacing
S.~Teerapittayanon, B.~McDanel, and H.~T. Kung, ``Branchynet: Fast inference via early exiting from deep neural networks,'' in \emph{23rd International Conference on Pattern Recognition, {ICPR} 2016, Canc{\'{u}}n, Mexico, December 4-8, 2016}.\hskip 1em plus 0.5em minus 0.4em\relax {IEEE}, 2016, pp. 2464--2469. [Online]. Available: \url{https://doi.org/10.1109/ICPR.2016.7900006}
\BIBentrySTDinterwordspacing

\bibitem{simulator1}
\BIBentryALTinterwordspacing
Y.~Liu, W.~Chen, Y.~Bai, X.~Liang, G.~Li, W.~Gao, and L.~Lin, ``Aligning cyber space with physical world: A comprehensive survey on embodied ai,'' 2024. [Online]. Available: \url{https://arxiv.org/abs/2407.06886}
\BIBentrySTDinterwordspacing

\bibitem{simulator2}
\BIBentryALTinterwordspacing
C.~Gao, B.~Zhao, W.~Zhang, J.~Mao, J.~Zhang, Z.~Zheng, F.~Man, J.~Fang, Z.~Zhou, J.~Cui, X.~Chen, and Y.~Li, ``Embodiedcity: A benchmark platform for embodied agent in real-world city environment,'' 2024. [Online]. Available: \url{https://arxiv.org/abs/2410.09604}
\BIBentrySTDinterwordspacing

\bibitem{simulator3}
\BIBentryALTinterwordspacing
J.~Gu, E.~Stefani, Q.~Wu, J.~Thomason, and X.~Wang, ``Vision-and-language navigation: A survey of tasks, methods, and future directions,'' in \emph{Proceedings of the 60th Annual Meeting of the Association for Computational Linguistics (Volume 1: Long Papers)}.\hskip 1em plus 0.5em minus 0.4em\relax Association for Computational Linguistics, 2022. [Online]. Available: \url{http://dx.doi.org/10.18653/v1/2022.acl-long.524}
\BIBentrySTDinterwordspacing

\bibitem{simulator4}
\BIBentryALTinterwordspacing
X.~Wang, D.~Yang, Z.~Wang, H.~Kwan, J.~Chen, W.~Wu, H.~Li, Y.~Liao, and S.~Liu, ``Towards realistic uav vision-language navigation: Platform, benchmark, and methodology,'' 2024. [Online]. Available: \url{https://arxiv.org/abs/2410.07087}
\BIBentrySTDinterwordspacing

\bibitem{simulator5}
P.~Anderson, Q.~Wu, D.~Teney, J.~Bruce, M.~Johnson, N.~S{\"u}nderhauf, I.~Reid, S.~Gould, and A.~van~den Hengel, ``Vision-and-language navigation: Interpreting visually-grounded navigation instructions in real environments,'' in \emph{Proceedings of the IEEE Conference on Computer Vision and Pattern Recognition (CVPR)}, 2018.

\bibitem{simulator6}
\BIBentryALTinterwordspacing
S.~Aldhaheri, Y.~Hu, Y.~Xie, P.~Wu, D.~Kanoulas, and Y.~Liu, ``Underwater robotic simulators review for autonomous system development,'' 2025. [Online]. Available: \url{https://arxiv.org/abs/2504.06245}
\BIBentrySTDinterwordspacing

\bibitem{simulator7}
\BIBentryALTinterwordspacing
X.~Wang, D.~Yang, Z.~Wang, H.~Kwan, J.~Chen, W.~Wu, H.~Li, Y.~Liao, and S.~Liu, ``Towards realistic uav vision-language navigation: Platform, benchmark, and methodology,'' 2024. [Online]. Available: \url{https://arxiv.org/abs/2410.07087}
\BIBentrySTDinterwordspacing

\end{thebibliography}

\begin{IEEEbiography}[{\includegraphics[width=1in,height=1.2in,clip,keepaspectratio]{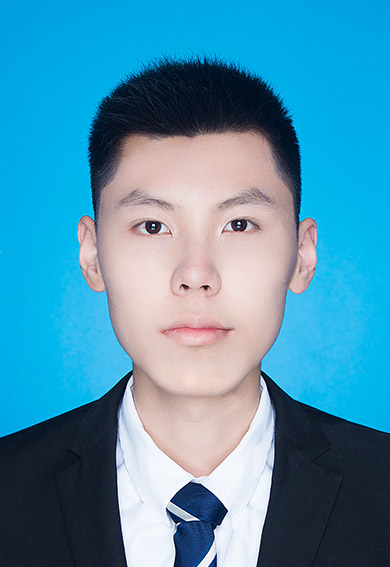}}]{Yida Xue} is currently a Ph.D. student at Zhejiang University, Hangzhou, China. He received his M.S. degree from Wuhan University, Wuhan, China. His research interests include knowledge graphs and natural language processing. 
\end{IEEEbiography}

\begin{IEEEbiography}[{\includegraphics[width=1in,height=1.2in,clip,keepaspectratio]{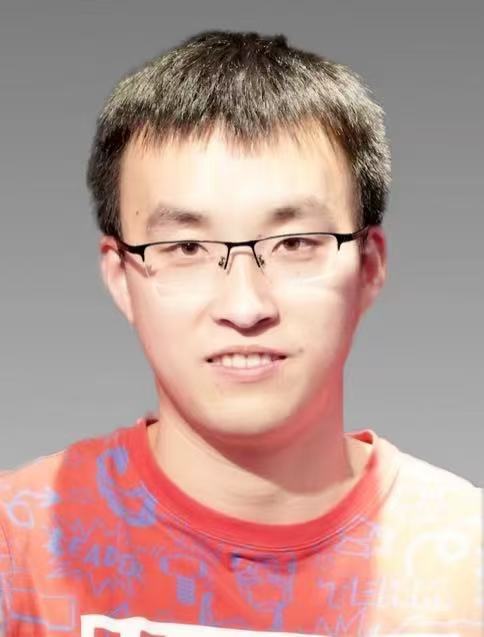}}]{Zhen Bi} received the B.S. degree from Northwestern Polytechnical University, Xian, China, in 2019, and he received the Ph.D. degree in Zhejiang University. His research interests are knowledge graph representation learning and natural language processing
\end{IEEEbiography}

\begin{IEEEbiography}[{\includegraphics[width=1in,height=1.2in,clip,keepaspectratio]{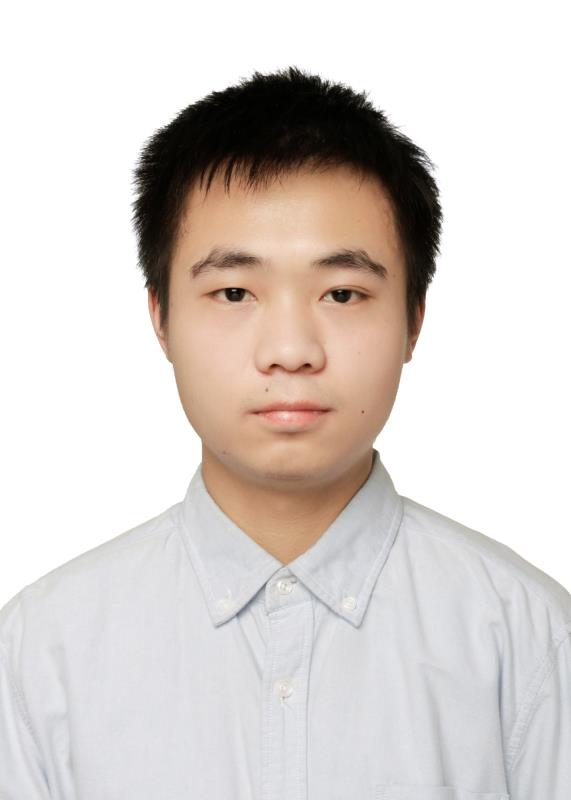}}]{Jinnan Yang} is currently a Ph.D. student at Nanjing University of Science and Technology, Nanjing, China.
He received the B.S. degree from Institute of Disaster Prevention, Langfang, China, in 2022, and he received the M.S. degree in Huzhou University, Huzhou, China. His research interests are computer vision and large language models.
\end{IEEEbiography}

\begin{IEEEbiography}[{\includegraphics[width=1in,height=1.2in,clip,keepaspectratio]{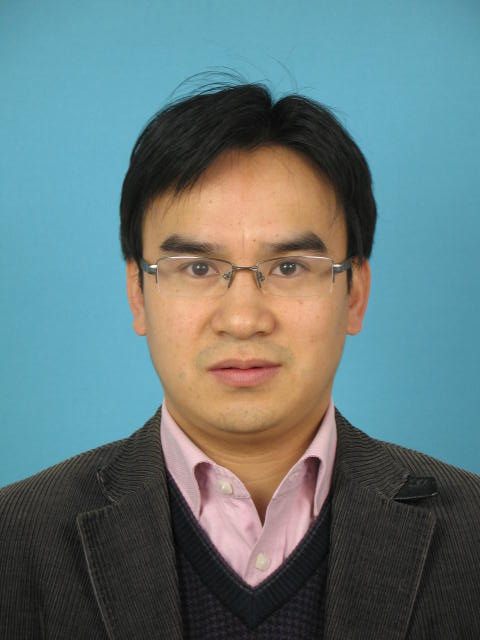}}]{Jungang Lou} received the Ph.D. degree in computer science and technology from Tongji University, Shanghai, China, in 2010. He is currently a Professor of Huzhou University, Huzhou, China. His current research interests include artificial intelligence, dependable computing and reliability engineering. He has published over 100 papers in refereed journals including IEEE TC, IEEE TCAD, IEEE TIFS, IEEE TNNLS, IEEE TCYB, IEEE TCSS and so on. 
\end{IEEEbiography}

\begin{IEEEbiography}[{\includegraphics[width=1in,height=1.25in,clip,keepaspectratio]{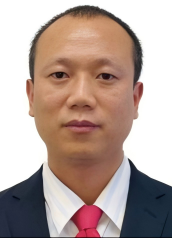}}]{Kehai Chen} (Member, IEEE) received the B.S. degree in computer science from the Xi’an University of Technology Xi’an, China, in 2010, the M.S. degree in computer science from the University of Chinese Academy of Sciences, Beijing, China, in 2013, and the Ph.D. degree in computer science from the Harbin Institute of Technology, Harbin, China, in 2018. He is a currently a Professor with the School of Computer Science and Technology, Harbin Institute of Technology (Shenzhen), Shenzhen, China, since 2023. He is also a Researcher with the National Institute of Information and Communications Technology, Tokyo, Japan, since 2018. His research interests include natural large language model, natural language processing, agents, multi-modal generation, and trustworthy reasoning.
\end{IEEEbiography}

\begin{IEEEbiography}[{\includegraphics[width=1in,height=1.25in,clip,keepaspectratio]{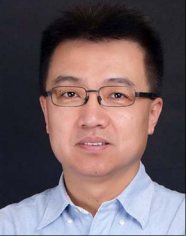}}]{Min Zhang} received the Ph.D. degree in computer
science from the Harbin Institute of Technology,
Harbin, China, in 1997. He is currently a Distinguished Professor with the School of Computer Science and Technology, Harbin Institute of Technology
(Shenzhen), Shenzhen, China, since 2021. He has
authored 150 papers in top-tier journals and conferences. His current research interests include machine
translation, natural language processing, and artificial
intelligence. He is the Vice President of COLIPS,
a Steering Committee Member of PACLIC, and an
Executive Member of AFNLP and ACL Fellow.
\end{IEEEbiography}

\begin{IEEEbiography}[{\includegraphics[width=1in,height=1.25in,clip,keepaspectratio]{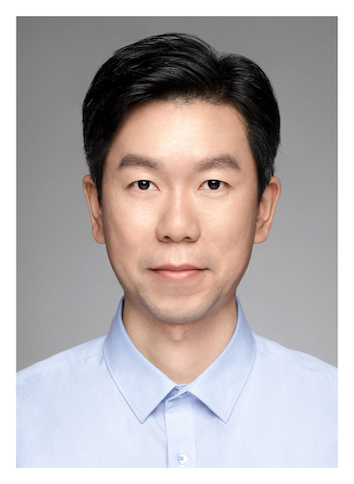}}]{Huajun Chen} is a full professor of College of Computer Science and Technologies at Zhejiang University, serve as the Director of Joint Lab on Knowledge Engine of AZFT (Alibaba-Zhejiang University Joint Research Institute of Frontier Technologies), and a deputy director of the Key Lab of Big Data Intelligence at Zhejiang Province.  He received his bachelor's degree and a Ph.D. from Zhejiang University in 2000 and 2004, respectively. He worked as a visiting assistant professor at Yale Center for Medical Informatics, Yale University (From June 2006 to June 2007), and a visiting scholar at the School of Computer Science of Carnegie Mellon University (From June 2007 to August 2008).
\end{IEEEbiography}
 
\begin{IEEEbiography}[{\includegraphics[width=1in,height=1.25in,clip,keepaspectratio]{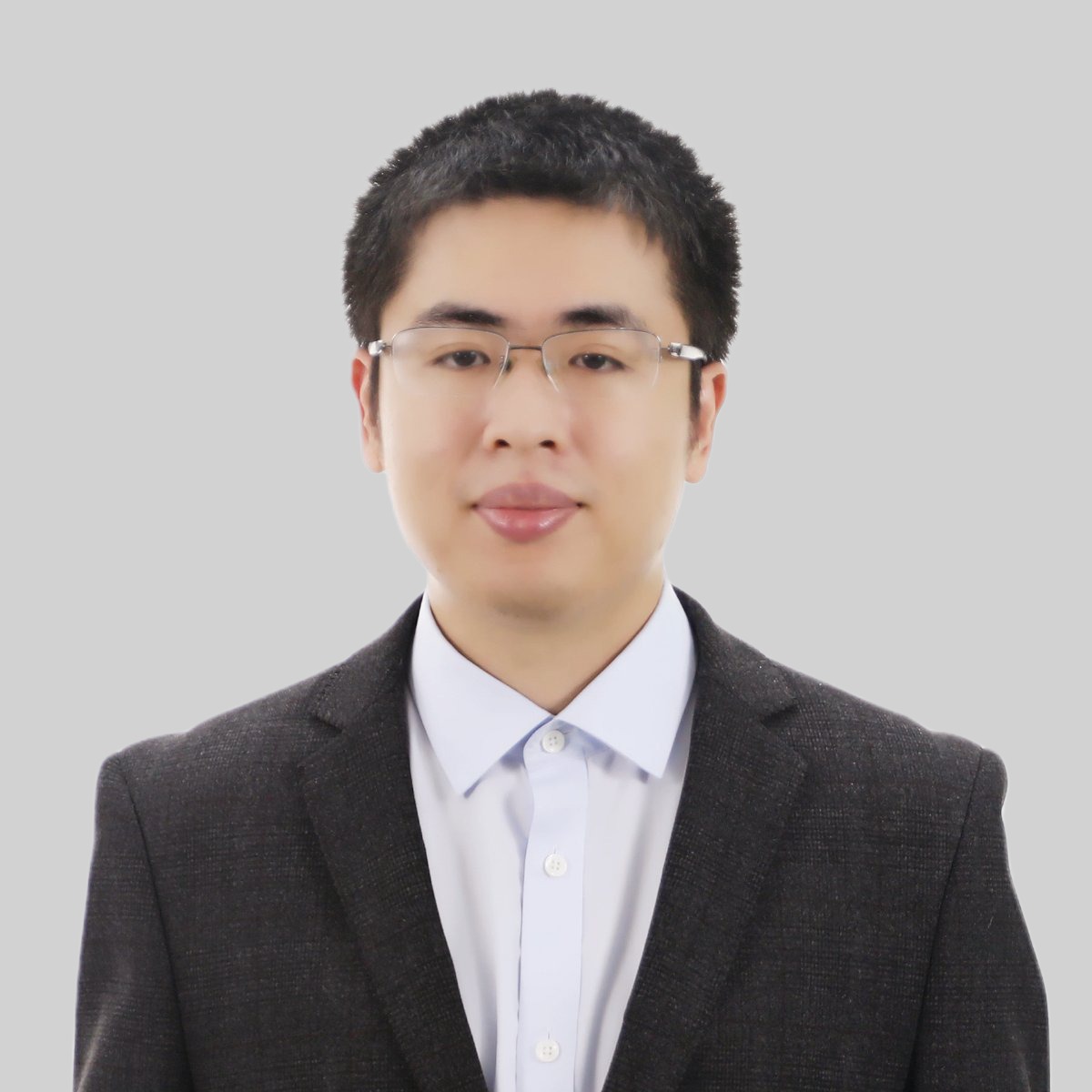}}]{Ningyu Zhang} (Member, IEEE) is an associate professor/doctoral supervisor at Zhejiang University, leading the group about KG and NLP technologies.
His research interests include natural language processing, information extraction, and large language models.
He has published many papers in top international academic conferences and journals such as Natural Machine Intelligence, Nature Communications, NeurIPS, ICLR, AAAI, IJCAI, WWW, KDD, SIGIR, ACL, ENNLP, NAACL, and IEEE/ACM Transactions on Audio Speech and Language. 
He has served as Area Chair for ACL 2023, ARR Action Editor, Senior Program Committee member for IJCAI 2023, Program Committee member for AAAI, NeurIPS, ICLR, WWW, SIGIR, KDD, ICML, AAAI, and reviewer for TKDE, TKDD. 
\end{IEEEbiography}
\end{document}